\documentclass[sigconf,noacm]{acmart}
\usepackage{multirow}
\usepackage{booktabs}
\usepackage{array}
\usepackage{microtype}
\usepackage{algorithm}
\usepackage{algpseudocode}
\usepackage{tabularx}
\usepackage{makecell}
\usepackage{array}
\usepackage{graphicx}
\usepackage{subcaption}
\usepackage{todonotes}

\renewcommand\footnotetextcopyrightpermission[1]{} 

\newcolumntype{Y}{>{\raggedright\arraybackslash}X}

\newcolumntype{L}[1]{>{\raggedright\arraybackslash}p{#1}}
\newcolumntype{R}[1]{>{\raggedleft\arraybackslash}p{#1}}
\AtBeginDocument{%
  }

\setcopyright{acmlicensed}
\copyrightyear{2026}
\acmYear{2026}
\acmDOI{XXXXXXX.XXXXXXX}
\acmISBN{978-1-4503-XXXX-X/2018/06}

\begin{document}

\title{Regression Models Meet Foundation Models: \\
A Hybrid-AI Approach to Practical Electricity Price Forecasting}


\author{Yunzhong Qiu}
\authornote{Both authors contributed equally to this research.}
\email{qiuyz24@mails.tsinghua.edu.cn}
\orcid{0009-0003-1034-1140}
\affiliation{%
  \institution{School of Software, BNRist \\ Tsinghua University}
  \city{Beijing}
  \country{China}
}

\author{Binzhu Li}
\email{lbz25@mails.tsinghua.edu.cn}
\orcid{0009-0006-9094-2131}
\authornotemark[1]
\affiliation{%
  \institution{School of Software, BNRist \\ Tsinghua University}
  \city{Beijing}
  \country{China}
}

\author{Hao Wei}
\email{weihao2@gcl-et.com}
\affiliation{%
  \institution{Suzhou~Industrial~Park~Xingzhixun~CS~Co., Ltd.}
  \city{Jiangsu}
  \country{China}
}

\author{Shenglin Weng}
\email{wengshenglin@gcl-et.com}
\affiliation{%
  \institution{Suzhou~Industrial~Park~Xingzhixun~CS~Co., Ltd.}
  \city{Jiangsu}
  \country{China}
}

\author{Chen Wang}
\email{wang_chen@tsinghua.edu.cn}
\affiliation{%
  \institution{School of Software, BNRist \\ Tsinghua University}
  \city{Beijing}
  \country{China}
}

\author{Zhongyi Pei}
\authornote{Corresponding Author.}
\email{peizhyi@tsinghua.edu.cn}
\affiliation{%
  \institution{School of Software, BNRist \\ Tsinghua University}
  \city{Beijing}
  \country{China}
}

\author{Mingsheng Long}
\email{mingsheng@tsinghua.edu.cn}
\affiliation{%
  \institution{School of Software, BNRist \\ Tsinghua University}
  \city{Beijing}
  \country{China}
}

\author{Jianmin Wang}
\email{jimwang@tsinghua.edu.cn}
\affiliation{%
  \institution{School of Software, BNRist \\ Tsinghua University}
  \city{Beijing}
  \country{China}
}

\renewcommand{\shortauthors}{Qiu et al.}

\begin{abstract}
  Electricity market prices exhibit extreme volatility, nonlinearity, and non-stationarity, making accurate forecasting a significant challenge. While cutting-edge time series foundation models (TSFMs) effectively capture temporal dependencies, they typically underutilize cross-variate correlations and non-periodic patterns that are essential for price forecasting. Conversely, regression models excel at capturing feature interactions but are limited to future-available inputs, ignoring crucial historical drivers that are unavailable at forecast time. To bridge this gap, we propose FutureBoosting, a novel paradigm that enhances regression-based forecasts by integrating forecasted features generated from a frozen TSFM. This approach leverages the TSFM’s ability to model historical patterns and injects these insights as enriched inputs into a downstream regression model. We instantiate this paradigm into a lightweight, plug-and-play framework for electricity price forecasting. Extensive evaluations on real-world electricity market data demonstrate that our framework consistently outperforms state-of-the-art TSFMs and regression baselines, achieving reductions in Mean Absolute Error (MAE) of more than 30\% at most. Through ablation studies and explainable AI (XAI) techniques, we validate the contribution of forecasted features and elucidate the model’s decision-making process. FutureBoosting establishes a robust, interpretable, and effective solution for practical market participation, offering a general framework for enhancing regression models with temporal context.
\end{abstract}



\maketitle
\section{Introduction}
A healthy energy market is essential for achieving high utilization of new energy and preventing global warming.
As a critical supporting technology, electricity price forecasting (EPF) plays a pivotal role in activating energy markets, by which market participants are able to make informed decisions, including bidding strategies, risk management, and resource allocation. 
However, electricity prices are inherently volatile, nonlinear, and non-stationary due to increasingly complex supply-demand dynamics, posing a significant challenge for accurate EPF with cutting-edge AI technologies.

\begin{figure}[htbp]
  \centering
  \begin{subfigure}[b]{\linewidth}
    \centering
    \includegraphics[width=\linewidth]{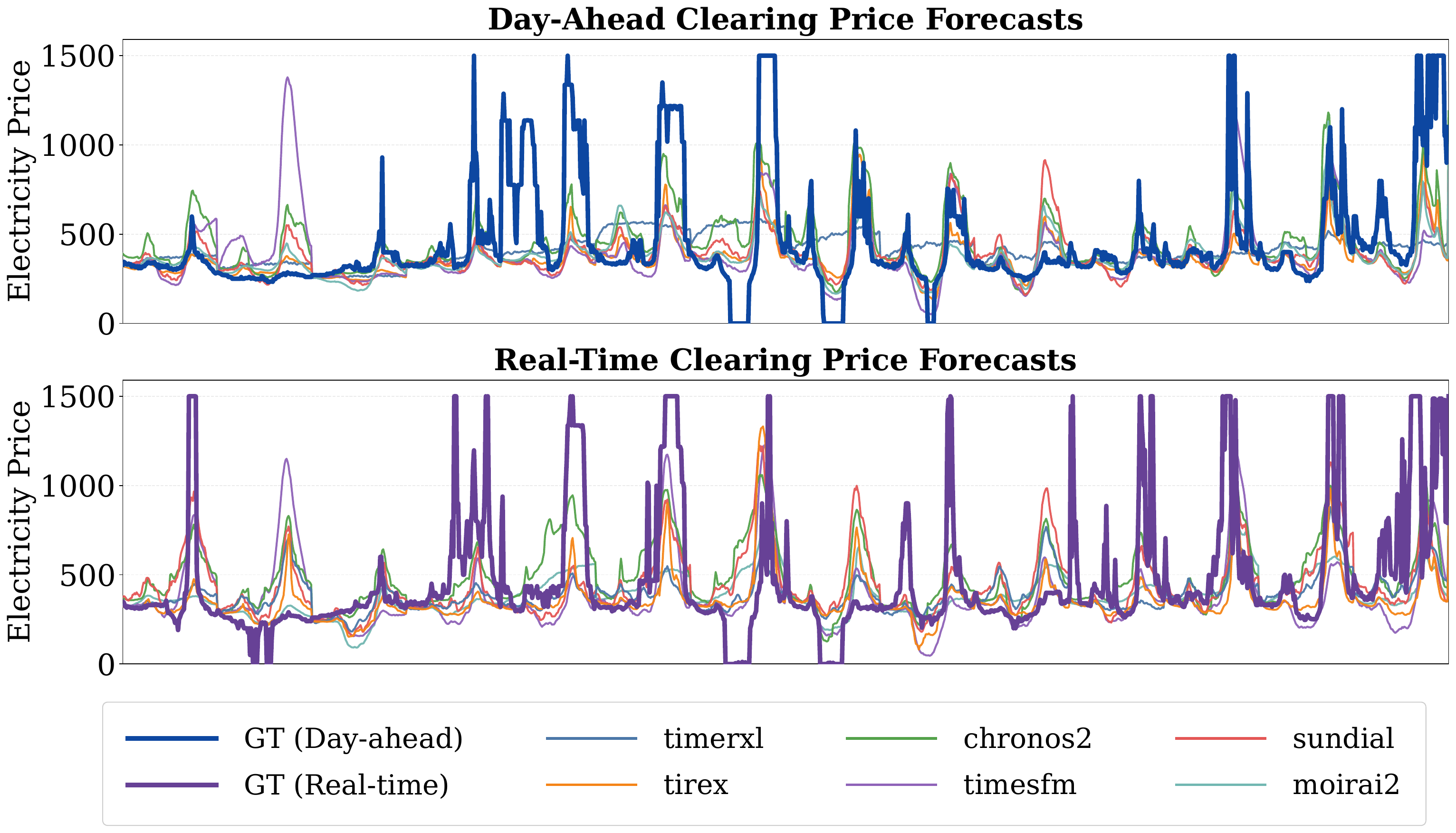}
    \caption{Volatile, nonlinear, and non-stationary temporal dynamics.}
    \label{fig:inference_comparison}
  \end{subfigure}

  \begin{subfigure}[b]{\linewidth}
    \centering
    \includegraphics[width=\linewidth]{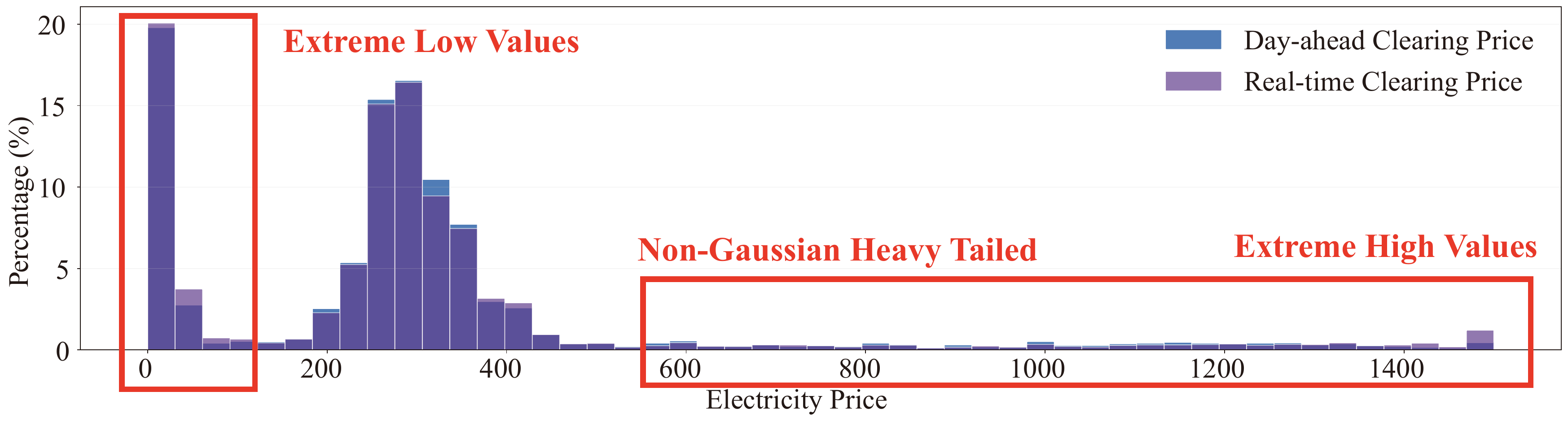}
    \caption{Non-Gaussian, heavy-tailed, and extreme value distribution.}
    \label{fig:price_distribution_comparison}
  \end{subfigure}
    \caption{
    Electricity price characteristics in the Shanxi spot market (15-min).
    (a) Day-ahead vs real-time forecasts in a representative October 2025 window.
    (b) Price distribution over the full year 2025, showing heavy tails and extremes.
    }

  \label{fig:shanxi_spot_data}
\end{figure}

Auto-regressive and regression modeling are two mainstream paradigms in electricity price forecasting.
Recently, time-series foundation models (TSFMs) have emerged as a state-of-the-art approach in the auto-regressive paradigm, thanks to their large-scale pretraining on diverse time series corpora.
On public benchmarks~\cite{aksu2024gift,shchur2025fev}, TSFMs ~\cite{DBLP:journals/corr/abs-2510-15821,DBLP:conf/iclr/LiuQ00L25,DBLP:conf/icml/LiuQSCY00L25,DBLP:journals/corr/abs-2511-11698,DBLP:journals/corr/abs-2505-23719,DBLP:conf/icml/DasKSZ24} demonstrate superior zero-shot capability, surpassing statistical models~\cite{Box1978TimeSA,gardner1985exponential} and some end-to-end deep learning models~\cite{Oreshkin2019NBEATSNB,Zeng2022AreTE,DBLP:conf/iclr/NieNSK23,DBLP:journals/corr/abs-2410-16928}.

However, as shown by Figure~\ref{fig:shanxi_spot_data}, these superior benchmark performance gains are not guaranteed in real-world electricity price forecasting, where electricity price exhibits volatile, nonlinear, and non-stationary temporal dynamics.
For electricity trading, the violent changes in prices have a significant impact on the supply-demand balancing and participants' profits.
By Table~\ref{tab:shanxi_difficulty_key_2025_nomismatch}, which summarizes key difficulty indicators computed over the full year of 2025, we find that both day-ahead and real-time prices exhibit a large tail ratio ($p99/p50\approx5$) and a non-negligible extreme-event frequency ($\mathbb{P}(p>1000)\approx6\%$).
The distribution is strongly non-Gaussian (excess kurtosis $\approx5$), and the change is also highly fierce (the top 1\% of absolute changes exceed 400; large jumps $|\Delta p|>200$ occur in about 3\% of intervals), implying that rare but impactful spikes are much more frequent than a normal assumption of noise.
These properties jointly make auto-regressive forecasting challenging. 


\begin{table}[t]
\centering
\caption{Key difficulty indicators of Shanxi electricity prices in 2025. Higher values indicate more challenging dynamics.}
\label{tab:shanxi_difficulty_key_2025_nomismatch}
\small
\setlength{\tabcolsep}{2pt}
\renewcommand{\arraystretch}{1.05}
\begin{tabular}{lccc}
\toprule
Indicator & Day-ahead & Real-time & Gaussian \\
\midrule
Tail ratio $p99/p50$ & 4.86 & 5.33 & $\approx$ 2 (1.5--2.5) \\
Extreme freq. $\mathbb{P}(p>1000)$ & 5.45\% & 5.88\% & $\approx$0\% ($<$0.1\%) \\
Excess kurtosis & 5.05 & 5.39 & 0 \\
Jump size $p99(|\Delta p|)$ & 399.8 & 439.9 & $\approx 200( 2.576 \times \sigma_{\Delta p})$ \\
Jump freq. $\mathbb{P}(|\Delta p|>200)$ & 2.92\% & 3.33\% & $\approx$1\% (0.5--1.5\%) \\
\bottomrule
\end{tabular}
\end{table}

Instead of relying on temporal dependencies like auto-regressive modeling, the regression modeling paradigm~\cite{sykes1993introduction} directly estimates correlations between features and demonstrates strong performance in previous literature~\cite {Saini2016ElectricityPF,Daz2019PredictionAE}.
However, correlations between variables are often diversified in different downstream domains,
making regression modeling hard to generalize across large-scale cross-domain data. 
As a representative branch in a variety of classical regression models \cite{Guyon2003AnIT,Kuhn2013AppliedPM,Domingos2012AFU,mcdonald2009ridge,ranstam2018lasso,hearst1998support,rigatti2017random,song2015decision}, gradient boosting trees~\cite{ke2017lightgbm,Chen2016XGBoostAS} demonstrate superior capability in handling structured feature interactions and capturing domain-specific patterns through their hierarchical decision structures.
However, traditional regression approaches are unable to capture temporal dynamics and are sensitive to data quality and feature engineering~\cite{Guyon2003AnIT,Kuhn2013AppliedPM,Domingos2012AFU}.
Such drawbacks limit the forward-looking predictive capability of regression models.

To address these limitations separately from auto-regressive and regression models, we propose a novel modeling paradigm, \emph{FutureBoosting}.
Our paradigm bridges these approaches through a two-stage process:
(1) Using TSFMs to generate forecasts of uncertain future price drivers that are not yet available at the time of forecasting;
(2) Integrating these forecasts with available future variables into an enriched feature set, on which a regression model, such as LightGBM~\cite{ke2017lightgbm}, is trained to forecast final electricity prices. 
By bridging large-scale pretrained foundation models and lightweight machine learning models, we fully leverage generalizable cross-domain knowledge for a specific downstream domain. 

Our \emph{FutureBoosting} paradigm builds on a key insight: while TSFMs can generally capture temporal patterns that follow consistent and generalizable trends, cross‑variate dependencies tend to be highly task‑specific and require more domain‑aware modeling. By expanding the feature set with forecasts of future price drivers, we reframe the forecasting problem from directly modeling complex temporal‑price relationships to learning more interpretable cross‑variate associations within an enriched feature space that incorporates forward‑looking expectations.

Following the \emph{FutureBoosting} paradigm, we propose a lightweight, plug-and-play, and flexible EPF framework, based on which extensive experiments and analysis are conducted on real-world electricity market datasets, demonstrating significant improvements in forecasting accuracy over both standalone TSFMs and traditional cross-variate regression models. The major contributions are summarized as follows:
\begin{itemize}
  \setlength{\itemsep}{-1pt}
    \item \textbf{Innovative paradigm:} We propose the \emph{FutureBoosting} paradigm, a novel hybrid approach based on insights into the differences between TSFMs and regression modeling.
    \item \textbf{EPF framework:} Following the \emph{FutureBoosting} paradigm, we propose a lightweight, plug-and-play EPF framework based on TSFMs and regression models.    
    \item \textbf{Empirical validation:} We conduct comprehensive experiments and analysis on real-world electricity market datasets, demonstrating significant improvements compared to standalone auto-regressive and regression models.
\end{itemize}

\section{Related Works}

\subsection{Time Series Foundation Models}

Time series foundation models (TSFMs) are large-scale models pretrained on diverse time series corpora and designed for zero- or few-shot forecasting across domains without task-specific training.

The most prevalent structure in TSFMs is the transformer-based architecture, ranging from univariate decoder-only models to multivariate encoder-only models.
Representative work includes Chronos \cite{Ansari2024ChronosLT}, Chronos-2~\cite{DBLP:journals/corr/abs-2510-15821},  Timer-XL~\cite{DBLP:conf/iclr/LiuQ00L25} and Sundial~\cite{DBLP:conf/icml/LiuQSCY00L25}, Moirai~2.0~\cite{DBLP:journals/corr/abs-2511-11698} and TimesFM2.5~\cite{DBLP:conf/icml/DasKSZ24}.
Besides, an xLSTM-based model TiRex~\cite{DBLP:journals/corr/abs-2505-23719} also generates outstanding performance.
These models are typically evaluated on public benchmarks such as GiftEval~\cite{aksu2024gift} and FEV~\cite{shchur2025fev}, where they substantially outperform classical statistical methods (e.g., AutoARIMA~\cite{Box1978TimeSA}, AutoETS~\cite{gardner1985exponential}) and many end-to-end deep learning baselines (e.g., N-BEATS~\cite{Oreshkin2019NBEATSNB}, DLinear~\cite{Zeng2022AreTE}, PatchTST~\cite{DBLP:conf/iclr/NieNSK23}, xLSTM-Mixer~\cite{DBLP:journals/corr/abs-2410-16928}).
Their strength lies in capturing long-range temporal dependencies and generalizing across series via large-scale pretraining, and they are often used in zero-shot mode without task-specific fine-tuning.
However, benchmark superiority does not always carry over to domain-specific applications, such as electricity price forecasting, where cross-variate structure, regime shifts, and heavy-tailed price distributions may require complementary modeling beyond purely auto-regressive forecasting.

\subsection{Regression Models}

Regression analysis provides a broad family of methods for forecasting a continuous target from a set of exogenous variables.
Classical approaches include ordinary least squares (OLS)~\cite{burton2021ols}, generalized linear models (GLMs)~\cite{nelder1972generalized}, and polynomial regression~\cite{heiberger2009polynomial}; regularized variants such as Ridge~\cite{mcdonald2009ridge} and Lasso~\cite{ranstam2018lasso} control overfitting and support feature selection.
Nonlinear and nonparametric methods range from support vector regression (SVR)~\cite{hearst1998support} to tree-based ensembles: random forests~\cite{rigatti2017random}, gradient boosting (e.g., XGBoost~\cite{Chen2016XGBoostAS}, LightGBM~\cite{ke2017lightgbm}), and related decision-tree methods~\cite{song2015decision}.
The performance of these models is often highly dependent on feature engineering and selection, as they are sensitive to the relevance and scale of inputs~\cite{Guyon2003AnIT,Kuhn2013AppliedPM,Domingos2012AFU}.
For structured tabular data with mixed covariate types and strong interaction effects, tree-based methods—especially gradient boosting—have repeatedly shown strong empirical performance and are widely used in industrial forecasting applications~\cite{Chen2016XGBoostAS,ke2017lightgbm}.
In electricity price forecasting, tree-based regressors are commonly used to capture nonlinear relationships between prices and exogenous variables such as load, electricity mix, and temporal indicators. Their interpretability (e.g., via feature importance) also supports deep domain analysis~\cite{Saini2016ElectricityPF,Daz2019PredictionAE}.

\subsection{Electricity Price Forecasting}

Electricity price forecasting (EPF) has been studied from statistical, machine learning, and hybrid perspectives~\cite{lago2021forecasting}.
Early work relied on ARIMA~\cite{boxTimeSeriesAnalysis2015}, GARCH-type models~\cite{bollerslevGeneralizedAutoregressiveConditional1986} for volatility, and regression on load and other drivers~\cite{Weron2014ElectricityPF}. Later, neural networks, support vector machines, and tree-based models were applied to capture nonlinearities and complex feature interactions~\cite{Saini2016ElectricityPF,Daz2019PredictionAE,lago2021forecasting}.
Recent efforts have incorporated deep learning methods, such as LSTMs~\cite{Lago2018ForecastingSE} and transformers~\cite{Li2023DeepLB}, to model prices and related variables.
EPF is typically framed as day-ahead or real-time forecasting, with submission deadlines and horizons dictated by market rules (e.g., day-ahead markets requiring forecasts by a fixed time for the next day)~\cite{Kirschen2018FundamentalsEM}.
Key challenges include the non-Gaussian, heavy-tailed price data distribution, strong temporal cross-variate dependencies, renewable penetration impact, supply-demand balance, and non-stationarity due to regulatory and structural changes~\cite{Weron2014ElectricityPF}.
Our work connects this literature by combining zero-shot TSFMs for forecasting future-unavailable and future-uncertain drivers with a tree-based regressor that models cross-variate relationships on an enriched feature set, thus addressing both temporal generalization and domain-specific correlation structure in a single framework.

\section{Preliminaries}

\subsection{Electricity Market Trading}
\begin{figure}[htbp]
  \centering
  \includegraphics[width=\linewidth]{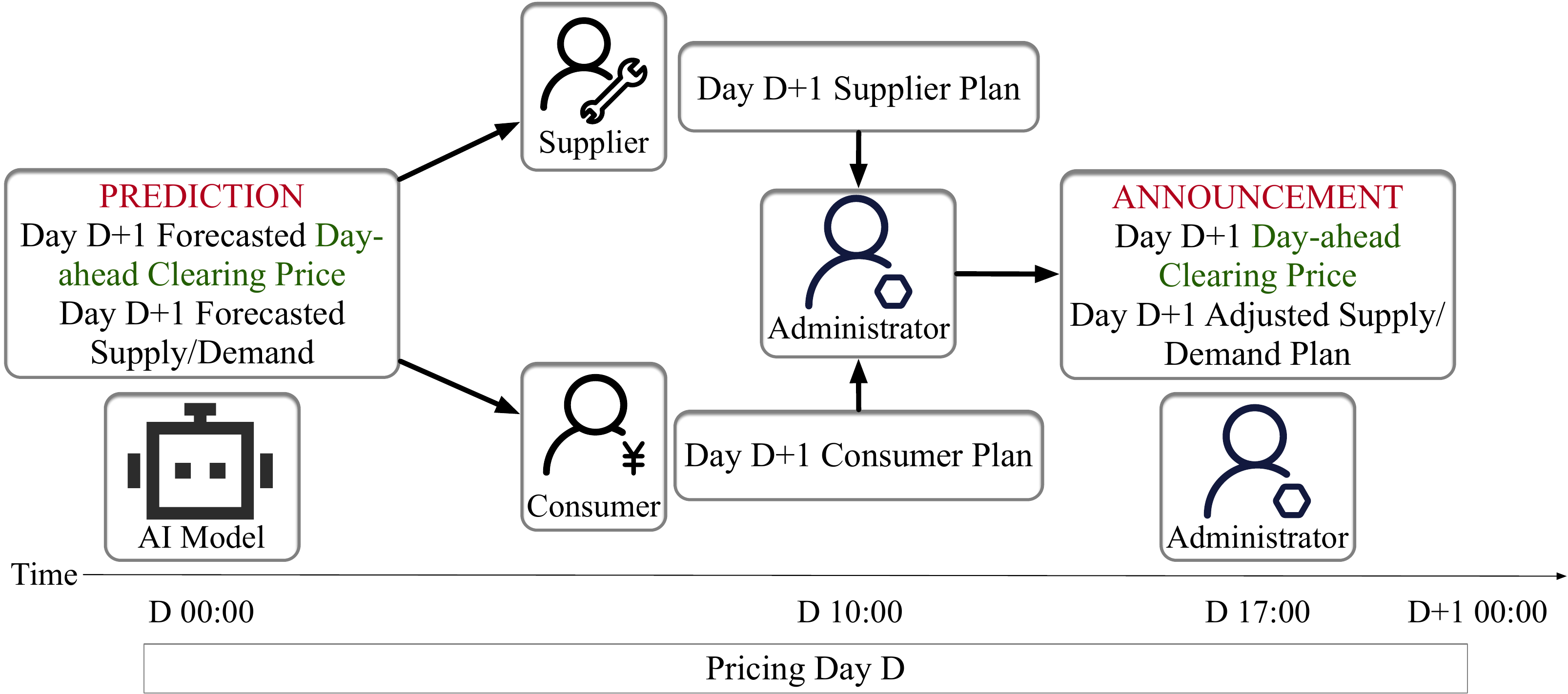}
  \caption{Illustration of electricity market trading background. Suppliers and consumers must provide day-ahead plans for day $D+1$ on day $D$. More accurate forecasts from the AI model improve the trading plan's foresight, yielding greater benefits.}
  \label{fig:background}
\end{figure}

The regulations in different electricity markets can vary in details, but they generally follow similar settings~\cite{haidar2021market,wang2024revisiting}.
Take the electricity market in China, for example. 
Electricity market prices are primarily divided into two parts: day-ahead and real-time prices.
As illustrated in Figure~\ref{fig:background}, before the trading process begins on Day $D+1$, the consumers and suppliers in the market are required to report their supply and demand plans for Day $D+1$ on Day $D$.
Then the trading platform, serving as an authoritative third party, aggregates all market participants' plans and derives the equitable day-ahead electricity prices for the Day $D+1$ market on Day $D$.

Since the day-ahead electricity market is planned in advance, actual supply and demand may not be fully scheduled. For example, establishments might adjust their electricity consumption in response to extreme weather or emerging events. The real-time electricity market is designed to support just-in-time electricity trading. Unlike the day-ahead electricity market price, which is set in advance, the real-time electricity market price is determined dynamically based on real-time supply and demand, and is more random and volatile.

For buyers and suppliers, both day-ahead and real-time electricity price forecasts are critical for developing trading strategies.
Additionally, an authoritative third party provides historical and future-available supply-and-demand data—including system load, transmission capacity, power generation, and interconnection plans—which are highly beneficial for electricity price forecasting.

\subsection{Problem Formulation}

The electricity prices and related grid data are published at 15-minute resolution, i.e., $H = 96$ timesteps per day.
We index each timestep within a day by $h \in \{1, \ldots, H\}$.

\textbf{Inputs.}
At the end of day $D$, three parts of data are accessible: $(\mathcal{Y}_{:D}, \mathbf{X}_{:D}, \mathbf{Z}_{D+1})$.

\emph{Historical target variable} ($\mathcal{Y}_{:D} \in \mathbb{R}^{D \cdot H \times 2}$): history day-ahead and real-time clearing prices for days $d \leq D$ and timesteps $h = 1,\ldots,H$.
\begin{equation}
\label{eq:input-history}
\mathcal{Y}_{:D} = \big\{ y_{d,h}^{\mathrm{da}}, \, y_{d,h}^{\mathrm{rt}} \;\big|\; d \leq D,\, h = 1,\ldots,H \big\}.
\end{equation}

\emph{Historical exogenous variables} ($\mathbf{X}_{:D} \in \mathbb{R}^{D \cdot H \times C}$): $C$ variables observed only up to day $D$, including historical load, power generation, and clearing prices of related markets.

\emph{Future available exogeneous variables} ($\mathbf{Z}_{D+1} \in \mathbb{R}^{H \times p}$): $p$ variables worked out in advance for the forecast day $D+1$, including timestamps, production plans, grid-provided forecasts, and weather forecasts for each timestep $\mathcal{M}$ on day $D+1$.

\textbf{Outputs.}
The goal is to forecast either the day-ahead or the real-time price curve for day $D+1$.
\begin{equation}
  \label{eq:output-future}
  \mathcal{Y}_{D+1} = \big\{ y_{d,h}^{\mathrm{da}}, \, y_{d,h}^{\mathrm{rt}} \;\big|\; D < d \leq D+1,\, h = 1,\ldots,H \big\}.
\end{equation}

\emph{Day-ahead price forecasting} ($f^{\mathrm{da}}$): the output is the vector of day-ahead clearing prices for day $D+1$. 
\begin{equation}
\label{eq:output-da}
\begin{aligned}
\hat{\mathbf{y}}_{D+1}^{\mathrm{da}} &= f^{\mathrm{da}}(\mathcal{Y}_{:D}, \mathbf{X}_{:D}, \mathbf{Z}_{D+1}), \\
\mathbf{y}_{D+1}^{\mathrm{da}} &= \big( y_{D+1,1}^{\mathrm{da}}, \ldots, y_{D+1,H}^{\mathrm{da}} \big)^\top \in \mathbb{R}^H.
\end{aligned}
\end{equation}

\emph{Real-time price forecasting} ($f^{\mathrm{rt}}$): the output is the vector of real-time clearing prices for day $D+1$. 
\begin{equation}
\label{eq:output-rt}
\begin{aligned}
\hat{\mathbf{y}}_{D+1}^{\mathrm{rt}} &= f^{\mathrm{rt}}(\mathcal{Y}_{:D}, \mathbf{X}_{:D}, \mathbf{Z}_{D+1}), \\
\mathbf{y}_{D+1}^{\mathrm{rt}} &= \big( y_{D+1,1}^{\mathrm{rt}}, \ldots, y_{D+1,H}^{\mathrm{rt}} \big)^\top \in \mathbb{R}^H.
\end{aligned}
\end{equation}

\section{Methodology}
\subsection{FutureBoosting Paradigm}
The electricity price in actual market scenarios is often driven by instantaneous supply-demand balance fluctuations, making the electricity price sensitive to relevant future time series expectations.
However, only historical observations and numerical weather prediction variables are available at planning time, making electricity price forecasting challenging due to information insufficiency.
Our proposed \emph{FutureBoosting} paradigm addresses the limitations in a two-stage pipeline:

\textbf{Stage 1 (forecast):} Given historical data $(\mathcal{Y}, \mathbf{X})$, we employ TSFMs to generate forecasts of future-unavailable variables denoted $\hat{\mathbf{X}}^{\mathrm{forecast}} = h(\mathcal{Y}, \mathbf{X})$, where $\mathcal{M}$ is the auto-regressive forecaster.
These forecasts serve as augmented features that capture temporal patterns and provide expectations of future supply-demand drivers.

\textbf{Stage 2 (regress):} We form an enriched feature set $\mathbf{F}$ that concatenates $\hat{\mathbf{X}}^{\mathrm{forecast}}$, future-available exogenous variables $\mathbf{Z}$, and optional domain-constructed factors. 
Then a regression model $f$ forecasts the target price $\hat{\mathbf{y}} = f(\mathbf{F})$.

\begin{figure}[htbp]
  \centering
  \includegraphics[width=\linewidth]{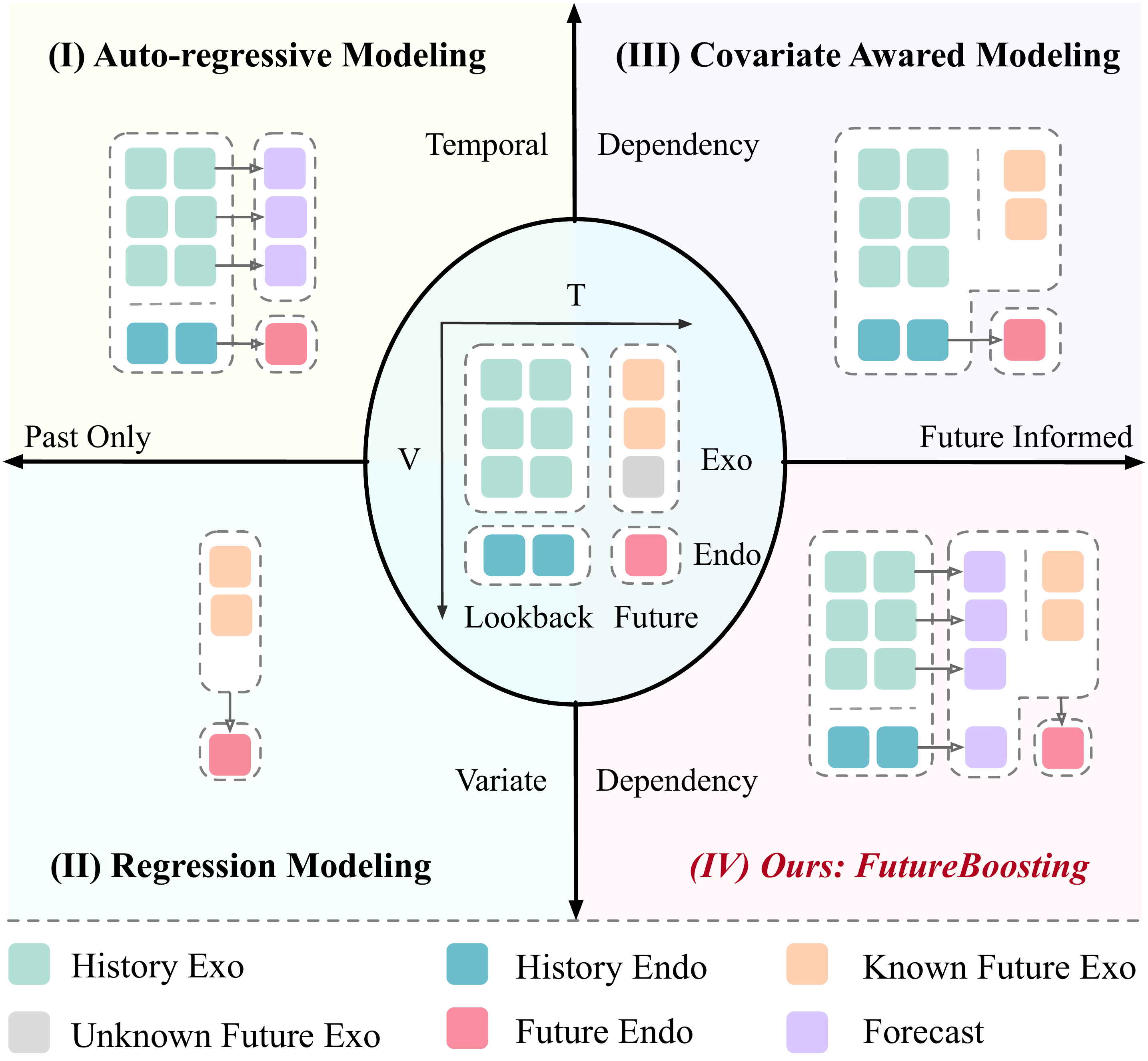}
  \caption{Comparison of forecasting paradigms. Our FutureBoosting paradigm first forecasts future-unavailable exogenous variables using TSFM, then uses these augmented features, along with future-available exogenous variables, for regression-based target forecasting.}
  \label{fig:paradigm}
\end{figure}

This design decouples temporal forecasting from cross-variate correlation modeling, enabling better leverage of the capabilities of TSFMs and regression models while compensating for each other's limitations.
Particularly, $\mathcal{M}$ provides high-fidelity forecasts of future drivers $\hat{\mathbf{X}}^{\mathrm{forecast}}$, and $f$ captures complex cross-variate relationships on the enriched feature space that includes these future expectations.
A comparison between the FutureBoosting paradigm and other existing paradigms is presented in Figure~\ref{fig:paradigm}.

\subsection{FutureBoosting Framework for EPF}
\begin{figure*}[t]
  \centering
  \includegraphics[width=\textwidth]{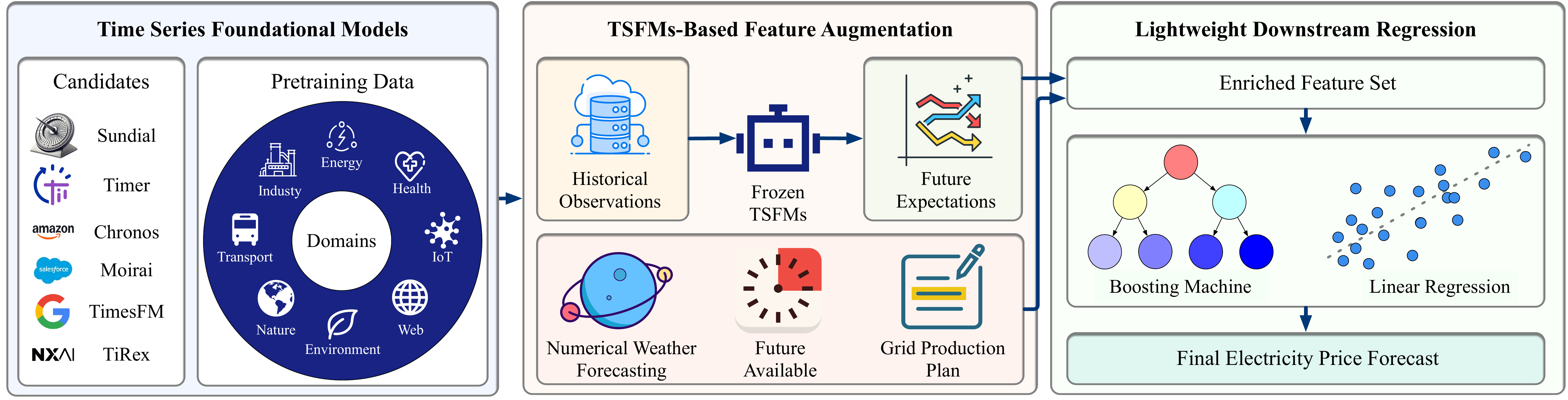}
  \caption{The pipeline of our FutureBoosting framework for electricity price forecasting. TSFM-augmented features, future available exogenous variables, and the grid production plan form an enriched feature set for downstream regression models.}
  \label{fig:main_figure}
\end{figure*}
\subsubsection{Framework Overview}
We instantiate the \emph{FutureBoosting} paradigm as a lightweight, plug-and-play framework for electricity price forecasting, which uses TSFM forecasts as a \emph{feature augmentation} mechanism to enrich the input space for downstream regression.
The framework operates in two sequential stages:


Given historical data $\mathcal{Y}_{:D}$ and $\mathbf{X}_{:D}$ available at the end of day $D$, we employ pre-trained TSFMs in zero-shot mode to generate forecasts for future-unavailable variables over the forecasting horizon for day $D+1$. These forecasts serve as augmented features that capture temporal patterns and provide expectations of future supply-demand drivers.
We then form an enriched feature set $\mathbf{F}_{D+1}$ by combining the forecasts, domain-knowledge-guided constructed factors (Section~\ref{sec:feature-pool}), and future available exogenous variables.

A tree-based regressor $f$ (e.g., LightGBM) is trained on the enriched feature set $\mathbf{F}_{D+1}$
to forecast electricity prices $\hat{\mathbf{y}}_{D+1}^{\mathrm{da}} = f(\mathbf{F}_{D+1})$ or $\hat{\mathbf{y}}_{D+1}^{\mathrm{rt}} = f(\mathbf{F}_{D+1})$, enabling the model to capture complex cross-variate interactions between TSFM-augmented features, domain-constructed factors, and future-available exogenous variables.

\subsubsection{TSFMs-Based Feature Augmentation}
\label{sec:feature}

We apply a pre-trained TSFM $\mathcal{M}$ to forecast a set of variables $\mathcal{V}=(\mathcal{Y}_{:D}, \mathbf{X}_{:D})$ over the horizon of day $D+1$. 
For each $v \in \mathcal{V}$, such as system load, renewable or thermal generation, $\mathcal{M}$ yields forecasts $\hat{v}_{D+1,1}, \ldots, \hat{v}_{D+1,H}$ for all $H$ timesteps. 
We assemble these into the augmented feature block,
which is then concatenated with domain-knowledge-guided constructed factors $\mathbf{C}_{D+1}$ and future available exogenous variables $\mathbf{Z}_{D+1}$ by Eq.~\ref{eq:D_plus_features}:
\begin{equation}
\label{eq:D_plus_features}
\begin{aligned}
\hat{\mathbf{X}}_{D+1}^{\mathrm{forecast}} & = \mathcal{M}(\mathcal{Y}_{:D}, \mathbf{X}_{:D}) \\
\mathbf{F}_{D+1} & = [\hat{\mathbf{X}}_{D+1}^{\mathrm{forecast}}, \mathbf{C}_{D+1}, \mathbf{Z}_{D+1}]
\end{aligned}
\end{equation}

The role of $\hat{\mathbf{X}}_{D+1}^{\mathrm{forecast}}$ is twofold: (i) it supplies expectations of future supply-demand drivers that are not directly observable at planning time, and (ii) it encodes long-range temporal structure via the TSFM's pretrained knowledge. 
The set $\mathcal{V}$ is selected based on domain knowledge: we include variables that are strongly correlated with electricity prices but are unavailable or uncertain at planning time (e.g., system load, renewable generation). Optionally, the TSFM can also forecast the target price series and include it as an augmented feature, allowing the downstream regressor to refine it using cross-variate information.

\subsubsection{Domain-knowledge-guided Constructed Factors}
\label{sec:feature-pool}

Beyond raw variables, we introduce domain-knowledge-guided constructed factors $\mathbf{C}_{D+1}$ that aggregate market structure information for the next trading day. These factors enable the downstream regressor to capture the supply-side structure more comprehensively by leveraging a richer feature pool.

\textbf{Thermal auction space} quantifies the effective capacity available for thermal generation in the day-ahead or real-time auction, reflecting the headroom for conventional units to set clearing prices when demand is high. 
It is computed as the difference between total thermal capacity and committed thermal generation, normalized by system load.

\textbf{Renewable ratio} is defined as the proportion of renewable generation (wind and solar) relative to total system load or generation.
It encodes the degree of renewable penetration, which strongly affects price levels and volatility due to near-zero marginal costs and intermittency. 
High renewable ratios typically correlate with lower average prices but higher price volatility.

\subsubsection{Lightweight Downstream Regression}
\label{sec:regression}

Given the enriched feature set $\mathbf{F}_{D+1}$, we train a regression model $f$ to forecast day-ahead or real-time electricity prices. In our Chinese electricity price forecasting experiments, we use a tree-based regressor, LightGBM~\cite{ke2017lightgbm}, to fit the enriched feature space and capture nonlinear interactions among TSFM-augmented features, domain-constructed factors, and future available exogenous variables.
The regressor is trained to minimize the mean squared error between predicted and observed prices. For a training set of $N$ samples, the loss is
\begin{equation}
\mathcal{L} = \frac{1}{N} \sum_{i=1}^{N} \left( y_i - \hat{y}_i \right)^2,
\end{equation}
where $y_i$ is the observed price and $\hat{y}_i = f(\mathbf{F}_{D+1,i})$ is the forecast. Standard regularization strategies, including early stopping and L2 regularization, are applied to reduce overfitting.
The trained model then yields forecasts $\hat{\mathbf{y}}_{D+1}^{\mathrm{da}}$ or $\hat{\mathbf{y}}_{D+1}^{\mathrm{rt}}$ for the target horizon. Explainability tools, such as feature importance and Shapley values, are used to analyze each feature's contribution to forecasts.

\subsection{Method Flexibility and Extensibility}
Take the above EPF framework as an example, the FutureBoosting-driven framework is intuitively flexible, enabling extension with various TSFMs and regression models.
Specifically, both univariate TSFMs (like TimerXL and Sundial) and multivariate TSFMs (like Chronos2) are supported.
For the regression model, any regressor, such as tree-based methods like XGBoost~\cite{Chen2016XGBoostAS} and Random Forest~\cite{rigatti2017random}, linear models~\cite{mcdonald2009ridge,ranstam2018lasso}, or other learners, can be applied in the FutureBoosting-driven framework. 
To demonstrate flexibility, we conduct comprehensive experiments using several state-of-the-art TSFMs on online data from China. 
Besides, additional experiments on public datasets are also provided in Section~\ref{sec:reale_results}, where another regression model is adopted.

\begin{table*}[t!]
\centering
\caption{China 12-month summary (Day-ahead on the left; Real-time on the right). For each metric: Val is the raw score; $\Delta$LGBM is the improvement vs LGBM (ic27); $\Delta$ZS is the improvement of FutureBoosting vs its corresponding zero-shot TSFM.}
\label{tab:shanxi_12m_dayahead_realtime_combo}
\small
\setlength{\tabcolsep}{2.6pt}
\renewcommand{\arraystretch}{1.10}

\begin{tabular*}{\textwidth}{@{\extracolsep{\fill}} l *{12}{c}}
\toprule
Method & \multicolumn{6}{c}{Day-ahead} & \multicolumn{6}{c}{Real-time} \\
\cmidrule(lr){2-7}\cmidrule(lr){8-13}
& \multicolumn{3}{c}{MSE} & \multicolumn{3}{c}{MAE} & \multicolumn{3}{c}{MSE} & \multicolumn{3}{c}{MAE} \\
\cmidrule(lr){2-4}\cmidrule(lr){5-7}\cmidrule(lr){8-10}\cmidrule(lr){11-13}
& Val & $\Delta$LGBM & $\Delta$ZS & Val & $\Delta$LGBM & $\Delta$ZS
& Val & $\Delta$LGBM & $\Delta$ZS & Val & $\Delta$LGBM & $\Delta$ZS \\
\midrule
LGBM (ic27) & 35658.52 & 0.00\% & -- & 100.53 & 0.00\% & -- & 49182.67 & 0.00\% & -- & 121.63 & 0.00\% & -- \\
\midrule

TimerXL & 76278.90 & -113.91\% & -- & 162.69 & -61.83\% & -- & 86353.72 & -75.58\% & -- & 171.63 & -41.11\% & -- \\
\textbf{ + FutureBoosting} & \textbf{33978.45} & \textbf{+4.71\%} & \textbf{+55.45\%} & \textbf{94.78} & \textbf{+5.72\%} & \textbf{+41.74\%} & \textbf{44522.21} & \textbf{+9.48\%} & \textbf{+48.44\%} & \textbf{117.10} & \textbf{+3.72\%} & \textbf{+31.77\%} \\
\cmidrule(lr){1-13}

Sundial & 62979.60 & -76.62\% & -- & 143.60 & -42.84\% & -- & 72219.22 & -46.84\% & -- & 154.30 & -26.86\% & -- \\
\textbf{ + FutureBoosting} & \textbf{35586.43} & \textbf{+0.20\%} & \textbf{+43.50\%} & \textbf{97.09} & \textbf{+3.42\%} & \textbf{+32.39\%} & \textbf{45415.52} & \textbf{+7.66\%} & \textbf{+37.11\%} & \textbf{117.45} & \textbf{+3.44\%} & \textbf{+23.88\%} \\
\cmidrule(lr){1-13}

Chronos2(mv) & 63465.48 & -77.98\% & -- & 144.60 & -43.84\% & -- & 72517.75 & -47.45\% & -- & 160.76 & -32.17\% & -- \\
\textbf{ + FutureBoosting} & \textbf{34344.18} & \textbf{+3.69\%} & \textbf{+45.89\%} & \textbf{94.90} & \textbf{+5.60\%} & \textbf{+34.37\%} & \textbf{46697.19} & \textbf{+5.05\%} & \textbf{+35.61\%} & \textbf{118.93} & \textbf{+2.22\%} & \textbf{+26.02\%} \\
\cmidrule(lr){1-13}

Chronos2(future) & 52197.23 & -46.38\% & -- & 137.00 & -36.27\% & -- & 64820.38 & -31.80\% & -- & 162.88 & -33.91\% & -- \\
\textbf{ + FutureBoosting} & \textbf{35315.01} & \textbf{+0.96\%} & \textbf{+32.34\%} & \textbf{97.14} & \textbf{+3.37\%} & \textbf{+29.09\%} & \textbf{46683.65} & \textbf{+5.08\%} & \textbf{+27.98\%} & \textbf{119.53} & \textbf{+1.73\%} & \textbf{+26.61\%} \\
\cmidrule(lr){1-13}

TimesFM & 66423.97 & -86.28\% & -- & 138.97 & -38.24\% & -- & 81627.66 & -65.97\% & -- & 154.98 & -27.42\% & -- \\
\textbf{ + FutureBoosting} & \textbf{35411.86} & \textbf{+0.69\%} & \textbf{+46.69\%} & \textbf{95.04} & \textbf{+5.46\%} & \textbf{+31.61\%} & \textbf{46778.28} & \textbf{+4.89\%} & \textbf{+42.69\%} & \textbf{118.04} & \textbf{+2.95\%} & \textbf{+23.84\%} \\
\cmidrule(lr){1-13}

Moirai2 & 65750.80 & -84.39\% & -- & 137.58 & -36.85\% & -- & 79852.29 & -62.36\% & -- & 155.49 & -27.84\% & -- \\
\textbf{ + FutureBoosting} & \textbf{34930.18} & \textbf{+2.04\%} & \textbf{+46.87\%} & \textbf{95.73} & \textbf{+4.77\%} & \textbf{+30.42\%} & \textbf{45987.65} & \textbf{+6.50\%} & \textbf{+42.41\%} & \textbf{117.70} & \textbf{+3.23\%} & \textbf{+24.30\%} \\
\cmidrule(lr){1-13}

TiRex & 67450.65 & -89.16\% & -- & 133.14 & -32.44\% & -- & 79122.98 & -60.88\% & -- & 143.93 & -18.33\% & -- \\
\textbf{ + FutureBoosting} & \textbf{35543.37} & \textbf{+0.32\%} & \textbf{+47.30\%} & \textbf{96.94} & \textbf{+3.57\%} & \textbf{+27.19\%} & \textbf{46944.86} & \textbf{+4.55\%} & \textbf{+40.67\%} & \textbf{117.35} & \textbf{+3.52\%} & \textbf{+18.47\%} \\
\midrule

\textbf{{Avg Improve}} &
-- & \textcolor{red}{\textbf{+1.80\%}} & \textcolor{red}{\textbf{+45.43\%}} &
-- & \textcolor{red}{\textbf{+4.56\%}} & \textcolor{red}{\textbf{+32.40\%}} &
-- & \textcolor{red}{\textbf{+6.17\%}} & \textcolor{red}{\textbf{+39.27\%}} &
-- & \textcolor{red}{\textbf{+2.97\%}} & \textcolor{red}{\textbf{+24.98\%}} \\
\bottomrule
\end{tabular*}
\end{table*}

\section{Experiments}
\label{sec:experiments}

\subsection{Experimental Settings}
\label{sec:exp_settings}

We introduce a real-world dataset \textbf{Shanxi} and a public benchmark \textbf{RealE} for the evaluation of electricity price forecasting.
\textbf{Shanxi} benchmark includes actual industrial electricity market data and weather conditions collected from Shanxi Province, China.
Two price objectives in the Shanxi dataset, namely the \emph{``day-ahead clearing price''} and the \emph{``real-time clearing price''}, are included in a monthly rolling evaluation from January to December in 2025.
\textbf{RealE} is a publicly presented benchmark in previous work~\cite{shao2025real} with sufficient spatial and temporal scope.
We use hourly data from France (FR) and Germany (DE) for evaluation, following the official split protocol.
Average \textbf{MSE} and \textbf{MAE} are reported for fair comparison.
Zero-shot TSFM inference and covariate-only regression are introduced as baselines.
We also fine-tune Chronos2~\cite{Ansari2024ChronosLT} following the official implementation as a representative baseline for fine-tuned TSFMs.
All inputs comply with the planning-time availability constraint to avoid leakage.
Dataset details, feature lists, baseline setups, and implementation are in Appendix~\ref{sec:appendix_exp_settings}.

\subsection{Forecasting Performance}
\label{sec:major_results}

\subsubsection{\textbf{Shanxi} Day-ahead and Real-time Results}
\label{sec:shanxi_12m}

Table~\ref{tab:shanxi_12m_dayahead_realtime_combo} summarizes the 12-month rolling evaluation on \textbf{Shanxi} for both day-ahead and real-time forecasting.
\emph{FutureBoosting} achieves consistent improvements
compared to zero-shot TSFM inference and regression models on exogenous variables.
Relative to zero-shot TSFMs,
\emph{FutureBoosting} reduces MSE/MAE by \textbf{45.43\%}/\textbf{32.40\%} for day-ahead and
\textbf{39.27\%}/\textbf{24.98\%} for real-time relative to the corresponding zero-shot TSFM baselines
($\Delta$ZS in Table~\ref{tab:shanxi_12m_dayahead_realtime_combo}).
Relative to regression models on exogenous variables, 
\emph{FutureBoosting} improves day-ahead MSE/MAE by \textbf{1.80\%}/\textbf{4.56\%} and real-time MSE/MAE by
\textbf{6.17\%}/\textbf{2.97\%} compared with LightGBM ($\Delta$LGBM), with larger gains on the more volatile real-time task.
Such results demonstrate the effectiveness of \emph{FutureBoosting} in practical electricity price forecasting.

Among all variants, TimerXL + FutureBoosting achieves the best overall performance, reaching 33978.45 MSE and 94.78 MAE for day-ahead,
and 44522.21 MSE and 117.10 MAE for real-time.
Month-wise results are provided in Table~\ref{tab:app_shanxi_12m_zs_hyb} (day-ahead) and Table~\ref{tab:app_shanxi_12m_rt_zs_hyb} (real-time) in the appendix.

\begin{table*}[htbp]
\centering
\caption{Real-E 24-step summary (FR on the left; DE on the right). For each metric: Val is the raw score; $\Delta$Linear is the improvement vs Regression-only Linear; $\Delta$ZS is the improvement of FutureBoosting vs its corresponding zero-shot TSFM.}
\label{tab:reale_fr_de_combo}
\small
\setlength{\tabcolsep}{2.6pt}
\renewcommand{\arraystretch}{1.10}

\begin{tabular*}{\textwidth}{@{\extracolsep{\fill}} l *{12}{c}}
\toprule
Method & \multicolumn{6}{c}{FR\_24} & \multicolumn{6}{c}{DE\_24} \\
\cmidrule(lr){2-7}\cmidrule(lr){8-13}
& \multicolumn{3}{c}{MSE} & \multicolumn{3}{c}{MAE} & \multicolumn{3}{c}{MSE} & \multicolumn{3}{c}{MAE} \\
\cmidrule(lr){2-4}\cmidrule(lr){5-7}\cmidrule(lr){8-10}\cmidrule(lr){11-13}
& Val & $\Delta$Linear & $\Delta$ZS & Val & $\Delta$Linear & $\Delta$ZS
& Val & $\Delta$Linear & $\Delta$ZS & Val & $\Delta$Linear & $\Delta$ZS \\
\midrule
Linear & 22.33 & 0.00\% & -- & 2.79 & 0.00\% & -- & 13.63 & 0.00\% & -- & 2.55 & 0.00\% & -- \\
\midrule

TimerXL & 4.41 & +80.25\% & -- & 1.37 & +50.86\% & -- & 3.00 & +77.99\% & -- & 1.21 & +52.48\% & -- \\
\textbf{ + FutureBoosting} & \textbf{3.66} & \textbf{+83.61\%} & \textbf{+16.98\%} & \textbf{1.26} & \textbf{+54.81\%} & \textbf{+7.72\%} & \textbf{1.85} & \textbf{+86.43\%} & \textbf{+38.40\%} & \textbf{0.95} & \textbf{+62.69\%} & \textbf{+21.36\%} \\
\cmidrule(lr){1-13}

Sundial & 4.00 & +82.09\% & -- & 1.29 & +53.73\% & -- & 4.23 & +68.96\% & -- & 1.42 & +44.23\% & -- \\
\textbf{ + FutureBoosting} & \textbf{3.45} & \textbf{+84.56\%} & \textbf{+13.75\%} & \textbf{1.20} & \textbf{+56.96\%} & \textbf{+6.74\%} & \textbf{2.32} & \textbf{+82.98\%} & \textbf{+45.12\%} & \textbf{1.09} & \textbf{+57.18\%} & \textbf{+23.25\%} \\
\cmidrule(lr){1-13}

Chronos2 & 3.61 & +83.83\% & -- & 1.23 & +55.89\% & -- & 2.28 & +83.28\% & -- & 1.03 & +59.55\% & -- \\
\textbf{ + FutureBoosting} & \textbf{2.90} & \textbf{+87.01\%} & \textbf{+19.53\%} & \textbf{1.11} & \textbf{+60.19\%} & \textbf{+9.63\%} & \textbf{1.49} & \textbf{+89.07\%} & \textbf{+34.55\%} & \textbf{0.87} & \textbf{+65.83\%} & \textbf{+15.69\%} \\
\cmidrule(lr){1-13}

TiRex & 3.76 & +83.16\% & -- & 1.24 & +55.53\% & -- & 2.31 & +83.06\% & -- & 1.05 & +58.76\% & -- \\
\textbf{ + FutureBoosting} & \textbf{3.15} & \textbf{+85.90\%} & \textbf{+16.24\%} & \textbf{1.14} & \textbf{+59.11\%} & \textbf{+8.58\%} & \textbf{1.63} & \textbf{+88.05\%} & \textbf{+29.51\%} & \textbf{0.90} & \textbf{+64.65\%} & \textbf{+13.87\%} \\
\cmidrule(lr){1-13}

Moirai2 & 3.81 & +82.94\% & -- & 1.25 & +55.17\% & -- & 2.30 & +83.13\% & -- & 1.07 & +57.96\% & -- \\
\textbf{ + FutureBoosting} & \textbf{3.21} & \textbf{+85.63\%} & \textbf{+15.63\%} & \textbf{1.16} & \textbf{+58.39\%} & \textbf{+7.86\%} & \textbf{1.53} & \textbf{+88.78\%} & \textbf{+33.77\%} & \textbf{0.88} & \textbf{+65.44\%} & \textbf{+17.16\%} \\
\cmidrule(lr){1-13}

TimesFM & 3.723 & +83.33\% & -- & 1.239 & +55.55\% & -- & 2.908 & +78.67\% & -- & 1.188 & +53.34\% & -- \\
\textbf{ + FutureBoosting} & \textbf{3.001} & \textbf{+86.56\%} & \textbf{+19.40\%} & \textbf{1.116} & \textbf{+59.98\%} & \textbf{+9.95\%} & \textbf{1.810} & \textbf{+86.72\%} & \textbf{+37.75\%} & \textbf{0.953} & \textbf{+62.57\%} & \textbf{+19.76\%} \\
\midrule
\textbf{Avg Improve} & -- &
 \textcolor{red}{\textbf{+85.54\%}} &  \textcolor{red}{\textbf{+16.92\%}} & -- &
 \textcolor{red}{\textbf{+58.24\%}} & \textcolor{red}{\textbf{+8.41\%}} & -- &
 \textcolor{red}{\textbf{+87.00\%}} & \textcolor{red}{\textbf{+36.52\%}} & -- &
 \textcolor{red}{\textbf{+63.06\%}} & \textcolor{red}{\textbf{+18.52\%}} \\
\bottomrule
\end{tabular*}
\end{table*}

\subsubsection{\textbf{RealE} France and Germany Results}
\label{sec:reale_results}

Table~\ref{tab:reale_fr_de_combo} reports results on RealE for France (FR) and Germany (DE).
Unlike \textbf{Shanxi}, zero-shot TSFMs generally outperform the covariate-only regressors on \textbf{RealE},
suggesting that temporal dependency is more dominant.

\emph{FutureBoosting} consistently brings additional improvements over zero-shot TSFMs and regression models on exogenous variables.
Relative to zero-shot TSFMs, \emph{FutureBoosting} improves MSE/MAE by \textbf{16.92\%}/\textbf{8.41\%} on FR and
\textbf{36.52\%}/\textbf{18.52\%} on DE ($\Delta$ZS).
Relative to regression models on exogenous variables, \emph{FutureBoosting} achieves substantial gains as well, improving MSE/MAE by
\textbf{85.54\%}/\textbf{58.24\%} on FR and \textbf{87.00\%}/\textbf{63.06\%} on DE ($\Delta$Linear).
Overall, Chronos2 + FutureBoosting performs best, achieving MSE/MAE of 2.90/1.11 on FR and 1.49/0.87 on DE.

\subsection{Ablation Studies}
\label{sec:ablations}

\subsubsection{Regressor Choices}
\label{sec:ablation_regressor}

Table~\ref{tab:regressor_choice} compares downstream regression models under the same TSFM backbone (Chronos2).
In \textbf{Shanxi}, LightGBM yields clear gains over linear regression for both day-ahead and real-time forecasting.
For the day-ahead electricity price forecast, MAE drops from 128.01 to 94.90, and MSE drops from 41165.74 to 34344.18.
For the real-time electricity price forecast, MAE drops from 142.88 to 118.93, and MSE drops from 50634.74 to 46697.19.
These consistent improvements on both metrics suggest a strongly nonlinear cross-variate mapping with interaction and threshold effects.

In \textbf{RealE}, the linear regression is consistently better than LightGBM.
For FR, MAE increases from 1.11 to 1.90, and MSE increases from 2.90 to 6.21.
For DE, MAE increases from 0.87 to 0.97, and MSE increases from 1.49 to 2.13.
This suggests that different data distributions exhibit different preferences for regression modeling.

\begin{table}[t]
\centering
\small
\caption{Regressor choice ablation with the same TSFM backbone (Chronos2).}
\label{tab:regressor_choice}
\setlength{\tabcolsep}{5pt}
\renewcommand{\arraystretch}{1.10}
\begin{tabular}{l cc cc c}
\toprule
Dataset & \multicolumn{2}{c}{Chronos2 + Linear} & \multicolumn{2}{c}{Chronos2 + LGBM} & Gain \\
\cmidrule(lr){2-3}\cmidrule(lr){4-5}
& MSE & MAE & MSE & MAE & (best) \\
\midrule
Shanxi (DA) & 41165.74 & 128.01 & \textbf{34344.18} & \textbf{94.90} & LGBM \\
Shanxi (RT) & 50634.74 & 142.88 & \textbf{46697.19} & \textbf{118.93} & LGBM \\
RealE (FR)  & \textbf{2.90} & \textbf{1.11} & 6.21 & 1.90 & Linear \\
RealE (DE)  & \textbf{1.49} & \textbf{0.87} & 2.13 & 0.97 & Linear \\
\bottomrule
\end{tabular}
\end{table}

\subsubsection{Fine-tuning Comparison}
\label{sec:ablation_finetune}

\begin{figure}[t]
  \centering
  \includegraphics[width=\linewidth]{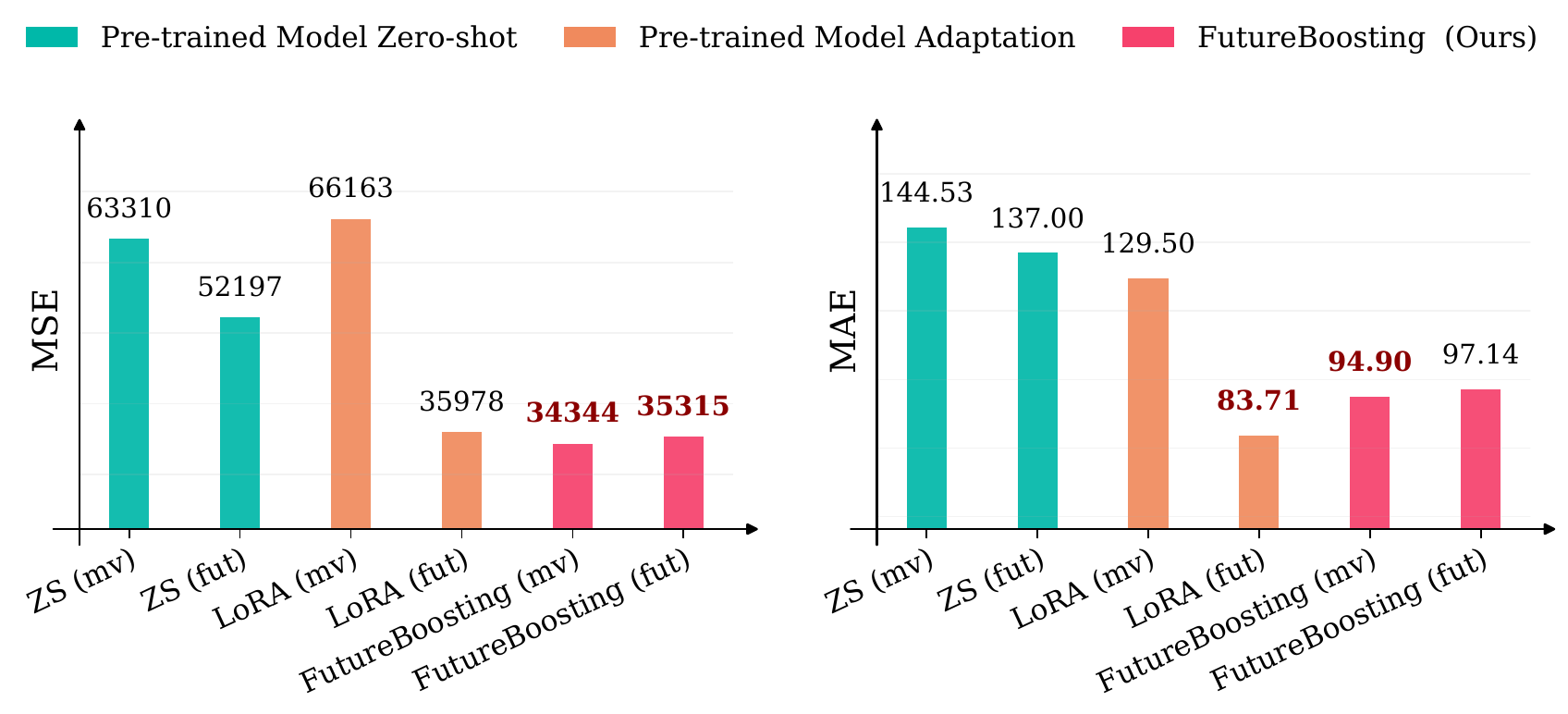}
  \caption{Adaptation study on Shanxi day-ahead forecasting. We compare zero-shot inference, LoRA fine-tuning, and FutureBoosting regression, with and without future-available exogenous variables.}
  \label{fig:chronos2_zs_lora_hybrid}
\end{figure}

Figure~\ref{fig:chronos2_zs_lora_hybrid} reports a fine-tuning comparison on Shanxi day-ahead forecasting (12-month average).
We select Chronos2, the latest state-of-the-art foundation model, as the backbone and compare three settings: zero-shot inference, LoRA fine-tuning, and \emph{FutureBoosting} with Chronos2 features.
Both multivariate inputs and future available exogenous variables are considered for the Chronos2 backbone.

\emph{FutureBoosting} substantially improves over Chronos2 zero-shot inference.
Under the multivariate setting, FutureBoosting reduces MSE/MAE from 63310.40/144.53 to 34344.18/94.90.
Adding future exogenous variables improves Chronos2 zero-shot inference (52197.23 MSE; 136.99 MAE).

Chronos2 LoRA fine-tuning with future exogenous variables achieves the lowest MAE (83.71).
This matches the MAE-based model selection used in our fine-tuning pipeline.
FutureBoosting achieves the lowest MSE (34344.18), indicating a metric trade-off between fine-tuning and regression in this setting.

\subsection{Efficiency Analysis}
\label{sec:efficiency}

\emph{FutureBoosting} is a lightweight framework that achieves performance comparable to finetuned TSFMs with much less computational resources.
We compare the efficiency performance of ``Chronos2 + FutureBoost'' and ``Chronos2+LoRA Fintuning'', results demonstrated in Table~\ref{tab:efficiency_compare}.
As \emph{FutureBoosting} apply zero-shot TSFMs inference strategy, we therefore further improve the inference efficiency through caching TSFM inference results.
This caching method makes the inference process of \emph{FutureBoosting} CPU-only, making it highly applicable to resource-limited computing devices.
More detailed efficiency analyses are presented in Appendix~\ref{sec:efficiency_detail}.

\begin{table}[t]
\centering
\small
\caption{Efficiency comparison on Shanxi day-ahead (per monthly window).}
\label{tab:efficiency_compare}
\setlength{\tabcolsep}{5pt}
\renewcommand{\arraystretch}{1.10}
\begin{tabular}{l c r r}
\toprule
Method & GPUs & GPU Mem (MB) & Total Time (s) \\
\midrule
FutureBoosting & 1$\times$4090 & 11049.85 & 151.57 \\
FutureBoosting (cached)   & CPU only & 0.00 & 59.24 \\
Fine-tuning  & 4$\times$4090 & 24518.00 & 8867.00 \\
\bottomrule
\end{tabular}
\end{table}

\section{Deployed System Evaluation}

\begin{figure}[htbp]
    \centering
        \includegraphics[width=\linewidth]{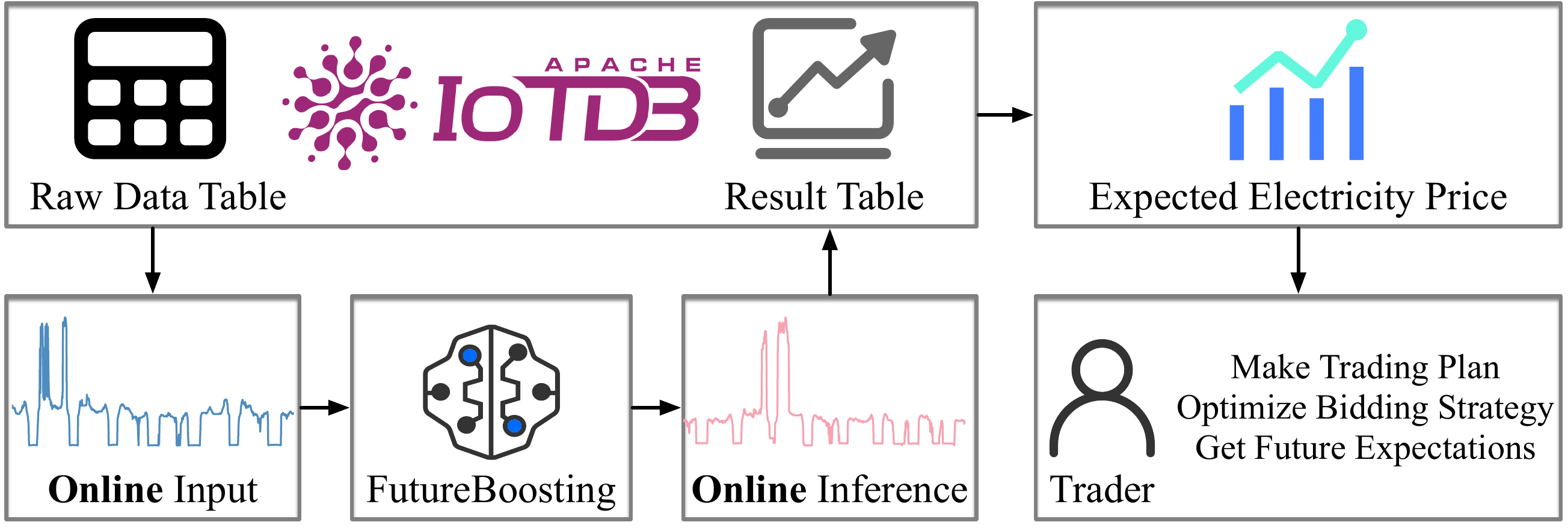}
    \caption{Application of FutureBoosting in Xingzhixun's IoTDB business server.}
    \label{fig:deploy}
\end{figure}

Our proposed solution, based on the \emph{FutureBoosting} paradigm, has been deployed on Xingzhixun's private IoTDB~\cite{wang2023apache} server for real-world online evaluation in December, 2025.
As illustrated in Figure~\ref{fig:deploy}, data relevant to electricity price forecasting are gathered and written into IoTDB every day.
For convenience, the raw data are reorganized and aligned from multiple sources into a single, timestamp-indexed time series table.
On each pricing day, the \emph{FutureBoosting} framework is invoked to infer the day-ahead clearing price and the real-time clearing price with the given IoTDB data series.

\begin{figure}[htbp]
  \centering
  \begin{subfigure}[t]{\linewidth}
    \centering
    \includegraphics[width=\linewidth]{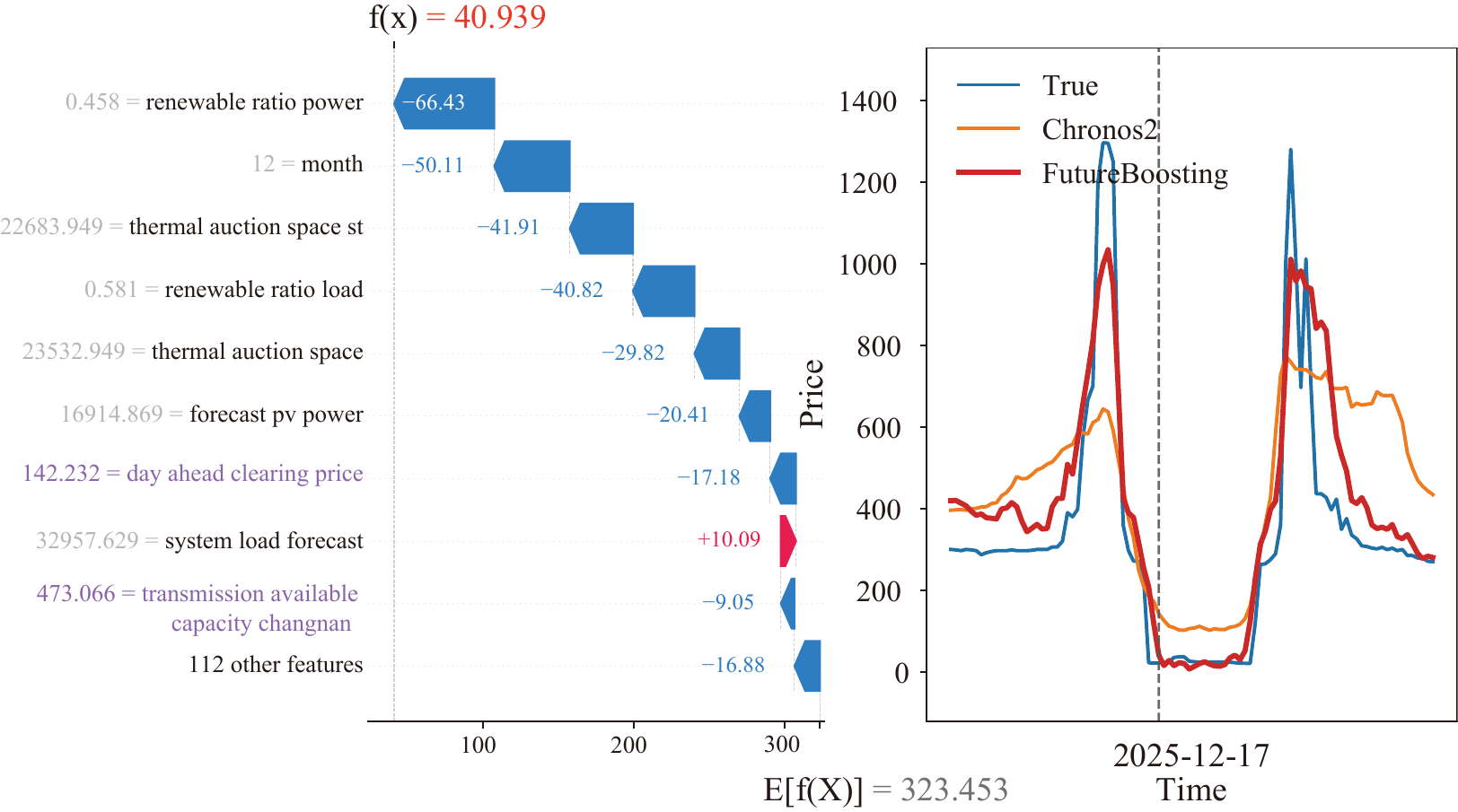}
    \caption{Low-price regime: FutureBoosting corrects the Chronos2 high bias and better follows the trough.}
    \label{fig:case_low}
  \end{subfigure}

  \begin{subfigure}[t]{\linewidth}
    \centering
    \includegraphics[width=\linewidth]{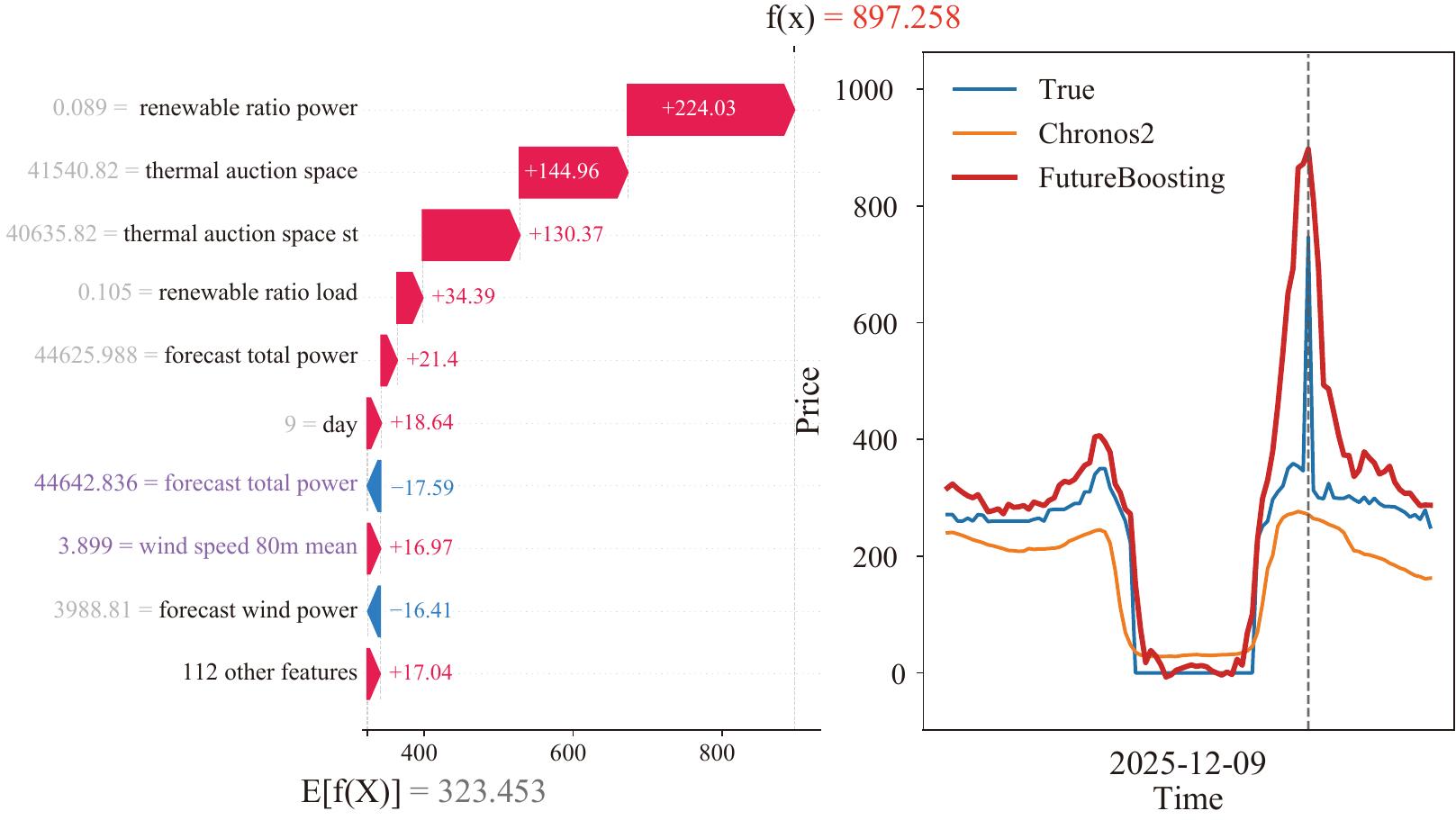}
    \caption{High-price regime: FutureBoosting amplifies the forecast and better matches the spike.}
    \label{fig:case_high}
  \end{subfigure}

  \caption{Sample SHAP explanations on deployed data. The TSFM-based feature derived from Chronos2 is highlighted.}
  \label{fig:shap_cases_low_high}
\end{figure}

For clarity, we include the Shapley value calculation in our analysis to demonstrate that our online deployment results are robust and explainable.
Figure~\ref{fig:shap_cases_low_high} demonstrates deployed showcases of the day-ahead clearing price in the Shanxi electricity market.
Compared with previous state-of-the-art time series foundational models such as Chronos2, \emph{FutureBoosting} outperforms in overall performance, especially for extreme electricity prices.
Specifically, Figure~\ref{fig:case_low} presents an example of a low-price regime on trading day December 17, 2025, and Figure~\ref{fig:case_high} presents an example of a high-price regime on trading day December 9, 2025.
Sample-level SHAP explanations are presented as waterfall plots, respectively.
The highlighted purple variables are TSFMs-based features generated by Chronos2, demonstrating the effectiveness of our enriched feature set design.
More detailed explainability assessments are presented in Appendix~\ref{sec:explainability_detail}.

\section{Conclusion and Future Works}
We propose FutureBoosting, a hybrid AI paradigm for electricity price forecasting that synergistically combines time-series foundation models (TSFMs) with regression. By using frozen TSFM forecasts as enriched forward-looking features, the method enables regression models to capture both complex temporal dynamics and cross-variate interactions. Evaluations on real-world market data show that FutureBoosting consistently outperforms standalone TSFMs and regression baselines. The framework is lightweight, interpretable via XAI, and offers a practical, high-accuracy solution for energy market decision-making.

Future directions include exploring automated feature selection methods to optimize the enriched feature set, particularly via TSFMs. Methods such as LASSO regularization or Shapley-based selection could help identify the most informative TSFM-derived forecasts and exogenous variables, reducing redundancy and enhancing model efficiency. Additionally, dynamic or adaptive feature selection strategies may be developed to respond to changing market regimes, further improving robustness and interpretability in real-world deployments.


\bibliographystyle{ACM-Reference-Format}
\bibliography{EPF}

\appendix

\section{FutureBoosting Framework}
\label{sec:appendix_framework}

Algorithm~\ref{alg:framework} formalizes the core logic of the FutureBoosting framework under the \emph{FutureBoosting} paradigm: \textbf{Stage 1} uses an auto-regressive model such as a state-of-the-art TSFM to forecast future-unavailable variables and form an enriched feature set; \textbf{Stage 2} trains a regression model on that feature set to predict day-ahead or real-time prices. 

\begin{algorithm}[htbp]
\caption{FutureBoosting Framework}
\label{alg:framework}
\begin{algorithmic}[1]
\Require Historical target $\mathcal{Y}_{:D}$, historical exogenous variables $\mathbf{X}_{:D}$, future-forecast exogenous variables $\mathbf{Z}_{D+1}$; TSFM (or auto-regressive forecaster) $\mathcal{M}$; variate list $\mathcal{V}$ to forecast; domain-constructed factors $\mathbf{C}_{D+1}$; regression model $f$
\Ensure Price forecasts $\hat{\mathbf{y}}_{D+1}^{\mathrm{da}}$ or $\hat{\mathbf{y}}_{D+1}^{\mathrm{rt}}$
\Statex \textbf{Stage 1: Auto-Regressive Feature Augmentation}
\For{each variate $v \in \mathcal{V}$}
  \State Run $\mathcal{M}$ in zero-shot mode on $\mathcal{V}(\mathcal{Y}_{:D}, \mathbf{X}_{:D})$ to produce forecasts $\hat{v}_{D+1,1}, \ldots, \hat{v}_{D+1,H}$ for day $D+1$
\EndFor
\State Assemble augmented features: $\hat{\mathbf{X}}_{D+1}^{\mathrm{forecast}} \gets \{\hat{v}_{D+1,\cdot}\}_{v \in \mathcal{V}}$
\State Form enriched feature set: $\mathbf{F}_{D+1} \gets [\hat{\mathbf{X}}_{D+1}^{\mathrm{forecast}}, \mathbf{C}_{D+1}, \mathbf{Z}_{D+1}]$
\Statex \textbf{Stage 2: Cross-Variate Regression}
\State Train regressor $f$ on $\mathbf{F}_{D+1}$ (and historical labels) to minimize forecasting error
\State Forecast: $\hat{\mathbf{y}}_{D+1}^{\mathrm{da}} \gets f(\mathbf{F}_{D+1})$ or $\hat{\mathbf{y}}_{D+1}^{\mathrm{rt}} \gets f(\mathbf{F}_{D+1})$
\State \Return $\hat{\mathbf{y}}_{D+1}^{\mathrm{da}}$ or $\hat{\mathbf{y}}_{D+1}^{\mathrm{rt}}$
\end{algorithmic}
\end{algorithm}

\section{Experimental Details}
\label{sec:appendix_exp_settings}

\subsubsection{Datasets}
\textbf{Shanxi Electricity Price Dataset.}
We use a real-world electricity market dataset from Shanxi, China, containing both
day-ahead and real-time clearing prices. The data is sampled every 15 minutes.
We perform D+1 forecasting at 00:00 each day with a 96-step horizon. The test period covers January--December 2025 with rolling monthly evaluation: for each test month, we train on the previous 12 months, validate on the previous 2 months, and test on the current month. 
To match real-world trading settings, we restrict Shanxi experiments to workdays and exclude dates that fall on Chinese statutory holidays.
Both day-ahead and real-time settings follow the same split. 
For direct regression on Shanxi, we use a selected set of 27 future-available exogenous variables (ic27), which include market tightness indicators, system operation and interconnection plans, renewable generation forecasts, and weather and calendar variables.
The complete covariate list is provided in Appendix~\ref{sec:appendix_features}.
For scaling, LightGBM~\cite{ke2017lightgbm} is trained on the original units of prices and exogenous variables, while TSFM inference follows each model's default preprocessing.
All reported metrics are computed on the original scale.

Table~\ref{tab:shanxi_difficulty_key_2025_nomismatch_appendix} summarizes key difficulty indicators computed over the full year of 2025.
In addition to the issues mentioned, the market is markedly non-stationary: the worst-case month-to-month distribution shift is large (KS$_\text{max}\approx0.48$), and the monthly high price level varies substantially (p95 range up to 1074).
These properties jointly make forecasting challenging, especially under rolling evaluation, where models must generalize across shifting regimes.

\begin{table}[t]
\centering
\caption{Key difficulty indicators of Shanxi electricity prices in 2025. Higher values indicate more challenging dynamics.}
\label{tab:shanxi_difficulty_key_2025_nomismatch_appendix}
\small
\setlength{\tabcolsep}{6pt}
\renewcommand{\arraystretch}{1.05}
\begin{tabular}{lrr}
\toprule
Indicator & Day-ahead & Real-time \\
\midrule
Tail ratio $p99/p50$ & 4.86 & 5.33 \\
Extreme freq. $\mathbb{P}(p>1000)$ & 5.45\% & 5.88\% \\
Excess kurtosis & 5.05 & 5.39 \\
Jump size $p99(|\Delta p|)$ & 399.8 & 439.9 \\
Jump freq. $\mathbb{P}(|\Delta p|>200)$ & 2.92\% & 3.33\% \\
Drift (KS$_\text{max}$, m/m$-$1) & 0.477 & 0.487 \\
Risk shift (monthly p95 range) & 763.9 & 1074.3 \\
\bottomrule
\end{tabular}
\end{table}

\textbf{RealE Benchmark Dataset.}
We use the RealE benchmark~\cite{shao2025real} and evaluate on the France (FR) and Germany (DE) subsets.
Both subsets are hourly, with the day-ahead price as the forecasting target.
We perform D+1 forecasting with a 24-hour horizon.
The FR subset spans January 2015 to June 2024, while the DE subset spans October 2018 to April 2023.
We split each subset into train/validation/test sets in a 0.7/0.1/0.2 ratio and merge train+validation when fitting regressors.
For direct regression, we use 15 exogenous variables for FR and 19 exogenous variables for DE; the full lists are provided in Appendix~\ref{sec:appendix_features}.
Following the benchmark protocol, RealE is standardized using mean and standard deviation computed on the training split only, and the same statistics are applied to validation and test splits to avoid leakage.
All RealE results are reported under this standardized evaluation setting.

\subsubsection{Covariate Feature Lists}
\label{sec:appendix_features}

In this section, we summarize the covariate sets used for \emph{direct regression} baselines and FutureBoosting regressors.
For Shanxi, we use a curated set of 27 future-available future exogenous variables (ic27). For RealE, we use 15 exogenous variables
(excluding the time index and the target), following the data availability in the benchmark.


The future-available exogenous variables (ic27) used in the Shanxi dataset are shown in Table~\ref{tab:ic27_features}.
The future-available exogenous variables used in the RealE dataset are shown in Table~\ref{tab:reale_features}.

\begin{table*}[htbp]
\centering
\small
\setlength{\tabcolsep}{6pt}
\renewcommand{\arraystretch}{1.15}
\caption{The set of 27 \textbf{future-available exogenous variables} (ic27) used as direct inputs to the regressor on the Shanxi dataset.}
\label{tab:ic27_features}
\begin{tabular}{L{3.6cm} L{13.0cm}}
\toprule
\textbf{Category} & \textbf{Variables} \\
\midrule
Market tightness \& constructed trading factors &
thermal\_auction\_space,
thermal\_auction\_space\_st,
renewable\_ratio\_load,
renewable\_ratio\_power \\
\midrule
System operation \& interconnection plans &
system\_load\_forecast,
day\_ahead\_interconnection\_plan\_total,
non\_market\_unit\_output,
transmission\_available\_capacity\_yanhuai \\
\midrule
Renewable / generation forecasts &
forecast\_total\_power,
forecast\_wind\_power,
forecast\_pv\_power,
forecast\_new\_energy\_total \\
\midrule
Calendar variables &
month,
weekday,
day \\
\midrule
Weather variables (summary statistics) &
dew\_point\_2m\_mean,
dew\_point\_2m\_max,
dew\_point\_2m\_min,
temperature\_2m\_mean,
temperature\_2m\_max,
relative\_humidity\_2m\_mean,
cloud\_cover\_low\_max,
precipitation\_probability\_mean,
extraterrestrial\_ghi\_mean,
surface\_pressure\_mean,
pressure\_msl\_mean,
pressure\_msl\_min \\
\bottomrule
\end{tabular}
\end{table*}


\begin{table*}[htbp]
\centering
\small
\setlength{\tabcolsep}{4pt}
\renewcommand{\arraystretch}{1.10}
\caption{Exogenous variables used for direct regression on RealE.}
\label{tab:reale_features}

\begin{tabularx}{\textwidth}{p{1.55cm} Y p{1.55cm} Y}
\toprule
\textbf{FR} & \textbf{Variables (15)} & \textbf{DE} & \textbf{Variables (19)} \\
\midrule

System / load &
\makecell[l]{%
genf\_Scheduled Generation\\
ntc\_Net Position\\
total\_Day-ahead Total Load Forecast\\
total\_Actual Total Load%
} &
System / load &
\makecell[l]{%
genf\_Scheduled Generation\\
ntc\_Net Position\\
total\_Actual Total Load%
} \\

\addlinespace[3pt]

Fuel-type generation &
\makecell[l]{%
Biomass -- Actual Aggregated\\
Fossil Gas -- Actual Aggregated\\
Fossil Hard coal -- Actual Aggregated\\
Fossil Oil -- Actual Aggregated\\
Hydro Pumped Storage -- Actual Aggregated\\
Hydro Run-of-river and poundage -- Actual Aggregated\\
Hydro Water Reservoir -- Actual Aggregated\\
Nuclear -- Actual Aggregated\\
Solar -- Actual Aggregated\\
Waste -- Actual Aggregated\\
Wind Onshore -- Actual Aggregated%
} &
Fuel-type generation &
\makecell[l]{%
Biomass -- Actual Aggregated\\
Fossil Brown coal/Lignite -- Actual Aggregated\\
Fossil Gas -- Actual Aggregated\\
Fossil Hard coal -- Actual Aggregated\\
Fossil Oil -- Actual Aggregated\\
Geothermal -- Actual Aggregated\\
Hydro Pumped Storage -- Actual Aggregated\\
Hydro Run-of-river and poundage -- Actual Aggregated\\
Hydro Water Reservoir -- Actual Aggregated\\
Nuclear -- Actual Aggregated\\
Other renewable -- Actual Aggregated\\
Solar -- Actual Aggregated\\
Waste -- Actual Aggregated\\
Wind Offshore -- Actual Aggregated\\
Wind Onshore -- Actual Aggregated\\
Other -- Actual Aggregated%
} \\

\bottomrule
\end{tabularx}
\end{table*}

\subsubsection{Evaluation Metrics}
We evaluate forecasting accuracy using Mean Squared Error (MSE) and Mean Absolute Error (MAE).
Given the ground-truth prices $\{y_t\}_{t=1}^{N}$ and forecasts $\{\hat{y}_t\}_{t=1}^{N}$, we compute:
\begin{equation}
\mathrm{MSE}=\frac{1}{N}\sum_{t=1}^{N}\left(y_t-\hat{y}_t\right)^2,
\end{equation}
\begin{equation}
\mathrm{MAE}=\frac{1}{N}\sum_{t=1}^{N}\left|y_t-\hat{y}_t\right|.
\end{equation}
For rolling evaluation on Shanxi, metrics are computed on each monthly test window and then averaged across the 12 months.

\subsubsection{Baselines}
\textbf{TSFM-only (zero-shot).}
We evaluate six pre-trained time-series foundation models (TSFMs) in a zero-shot manner:
Sundial~\cite{DBLP:conf/icml/LiuQSCY00L25}, TimerXL~\cite{DBLP:conf/iclr/LiuQ00L25}, Chronos2~\cite{DBLP:journals/corr/abs-2510-15821}, TiRex~\cite{DBLP:journals/corr/abs-2505-23719},
Moirai2~\cite{DBLP:journals/corr/abs-2511-11698}, and TimesFM~\cite{DBLP:conf/icml/DasKSZ24}. All TSFMs forcast the day-ahead/real-time price
directly from historical context without any downstream fine-tuning.
Chronos2 is used in a multivariate setting with two variants:
(i) Chronos2(mv), which performs multivariate inference using only historical exogenous variables and the target;
and (ii) Chronos2(future), which additionally takes the future-available values of exogenous variables as future inputs during inference.

\textbf{TSFM fine-tuning (LoRA).}
To benchmark task adaptation, we include fine-tuned TSFM baselines based on Chronos2 using LoRA.
We evaluate two fine-tuning settings: 
(i) Chronos2-FT(mv), which fine-tunes in a multivariate setting using historical exogenous variables and the target;
and (ii) Chronos2-FT(future), which additionally provides the future-available future exogenous variables as future inputs during training and inference.
All fine-tuned models are evaluated by directly forecasting prices (without any downstream regressor).

\textbf{Regressor-only (direct regression).}
We include a strong direct-regression baseline that forecasts prices solely from the future-available exogenous variables
(without TSFM forecasts). On the Shanxi dataset, we use LightGBM~\cite{ke2017lightgbm} with the selected 27 exogenous variables (ic27).
On RealE, we use Ridge regression with 15 exogenous variables for FR and 19 exogenous variables for DE.

\subsubsection{Implementation Details}
\label{sec:impl_details}
TSFM inference is performed on a single NVIDIA RTX 4090 GPU, while the downstream regressor is trained on CPU.

\textbf{TSFM inference.}
We perform zero-shot inference with a D+1 forecasting horizon.
Each TSFM follows its default sampling setup for probabilistic forecasting, with a batch size of 16. We cache TSFM forecasts to disk to avoid repeated inference across runs.
For the input context, we use model-specific historical window lengths in our implementation:
2880 for Sundial, 2048 for Chronos2 and TimesFM, and 1440 for the remaining TSFMs (including TimerXL, TiRex, and Moirai2).

\textbf{Regressor training.}
In Shanxi, we use LightGBM with 30,000 boosting rounds and 1,000 early-stopping rounds.
Key hyperparameters are: learning rate 0.05, num\_leaves 63, feature\_fraction 0.9,
bagging\_fraction 0.8, bagging\_freq 5, and min\_gain\_to\_split 0.08,
with 32 CPU threads and random seed 42.
In RealE, we use Ridge regression with default hyperparameters from scikit-learn~\cite{pedregosa2011scikit}.

\section{Detailed Efficiency Analysis}
\label{sec:efficiency_detail}

We report end-to-end efficiency on Shanxi day-ahead forecasting at the monthly-window level.
Our FutureBoosting pipeline instantiates Chronos2 inference once per window, then trains a lightweight regressor and evaluates on the same window.
To contextualize the computational cost, we compare FutureBoosting against LoRA fine-tuning under the same monthly setting and summarize the results in Table~\ref{tab:efficiency_compare}.

Without caching, TSFM inference takes 96.37s with a peak GPU memory allocation of 11,049.85~MB (peak reserved 12,142~MB).
The remaining stages are comparatively lightweight: feature construction takes 2.12s, LightGBM training takes 29.04s, and test-time evaluation takes 0.90s.
Optional SHAP analysis adds 23.14s.
Overall, the measured end-to-end runtime for one month is 151.57s, with a peak process CPU RSS of 2,058~MB.

With cached TSFM forecasts enabled, the same month reduces the inference stage to 4.24s and eliminates GPU usage in that step (0~MB peak allocation), while keeping the downstream cost similar:
feature construction 2.17s, LightGBM training 29.23s, evaluation 0.72s, and SHAP 22.88s.
This yields a total runtime of 59.24s per month, with a peak process CPU RSS of 1,295~MB.

In contrast, LoRA fine-tuning requires multi-GPU training.
For the same monthly window, our LoRA setting uses 4$\times$ RTX~4090 and takes approximately 2h~27m~47s end-to-end, with a monitored peak GPU memory allocation of 24,518~MB (Table~\ref{tab:efficiency_compare}).
Overall, fine-tuning incurs substantially higher wall-clock time and GPU resource demands, whereas FutureBoosting keeps the expensive TSFM inference cacheable and shifts most computation to efficient regression and evaluation.

\section{Detailed Explainability Assessments}
\label{sec:explainability_detail}

We assess interpretability for Shanxi day-ahead electricity price forecasting by applying SHAP to the downstream regressor in FutureBoosting.
All explanations are computed on the final LightGBM model trained on the enriched feature space, which combines TSFM rollouts with future-available exogenous variables.
We report both global feature importance and instance-level attributions.

\begin{figure}[htbp]
    \centering
        \includegraphics[width=\linewidth]{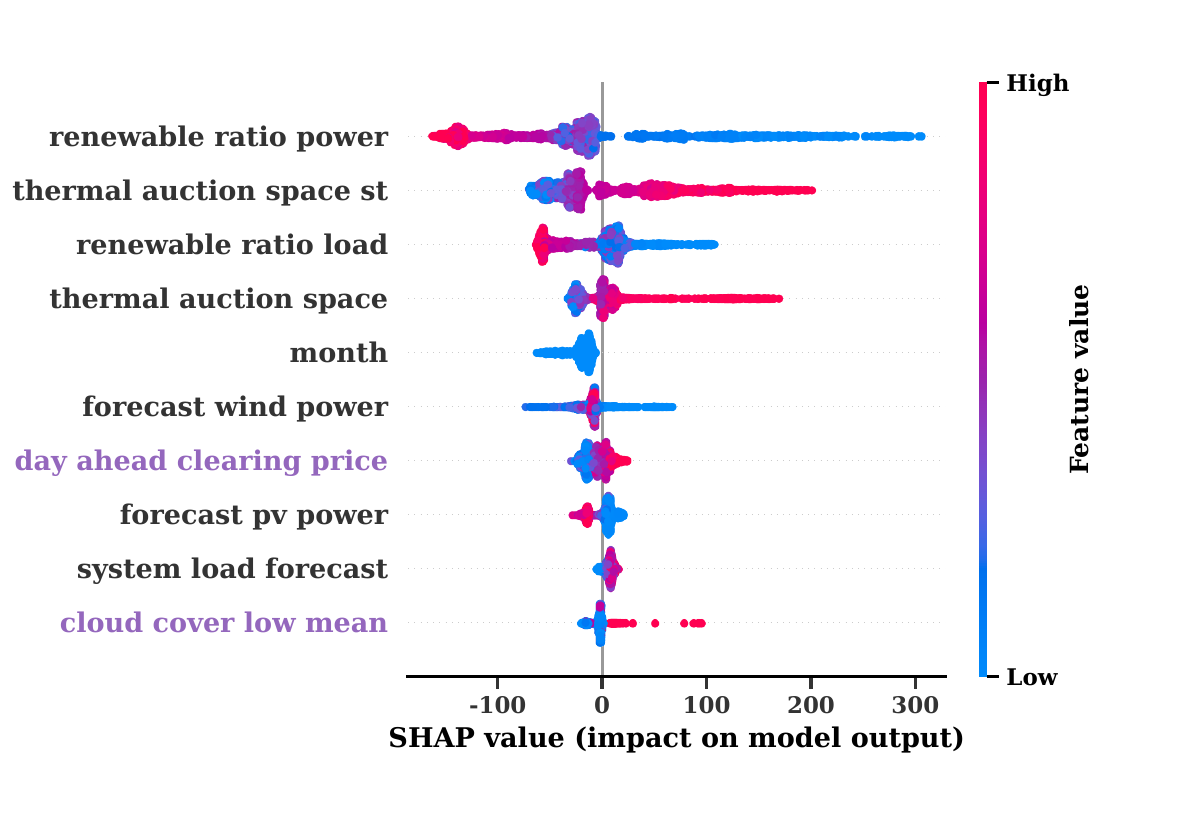}
    \caption{Global SHAP explanations on the test split. The Chronos2-derived forecast feature is highlighted to distinguish TSFM-predicted drivers from other exogenous variables.}
    \label{fig:shap_global}
\end{figure}

\begin{figure}[htbp]
    \centering
        \includegraphics[width=\linewidth]{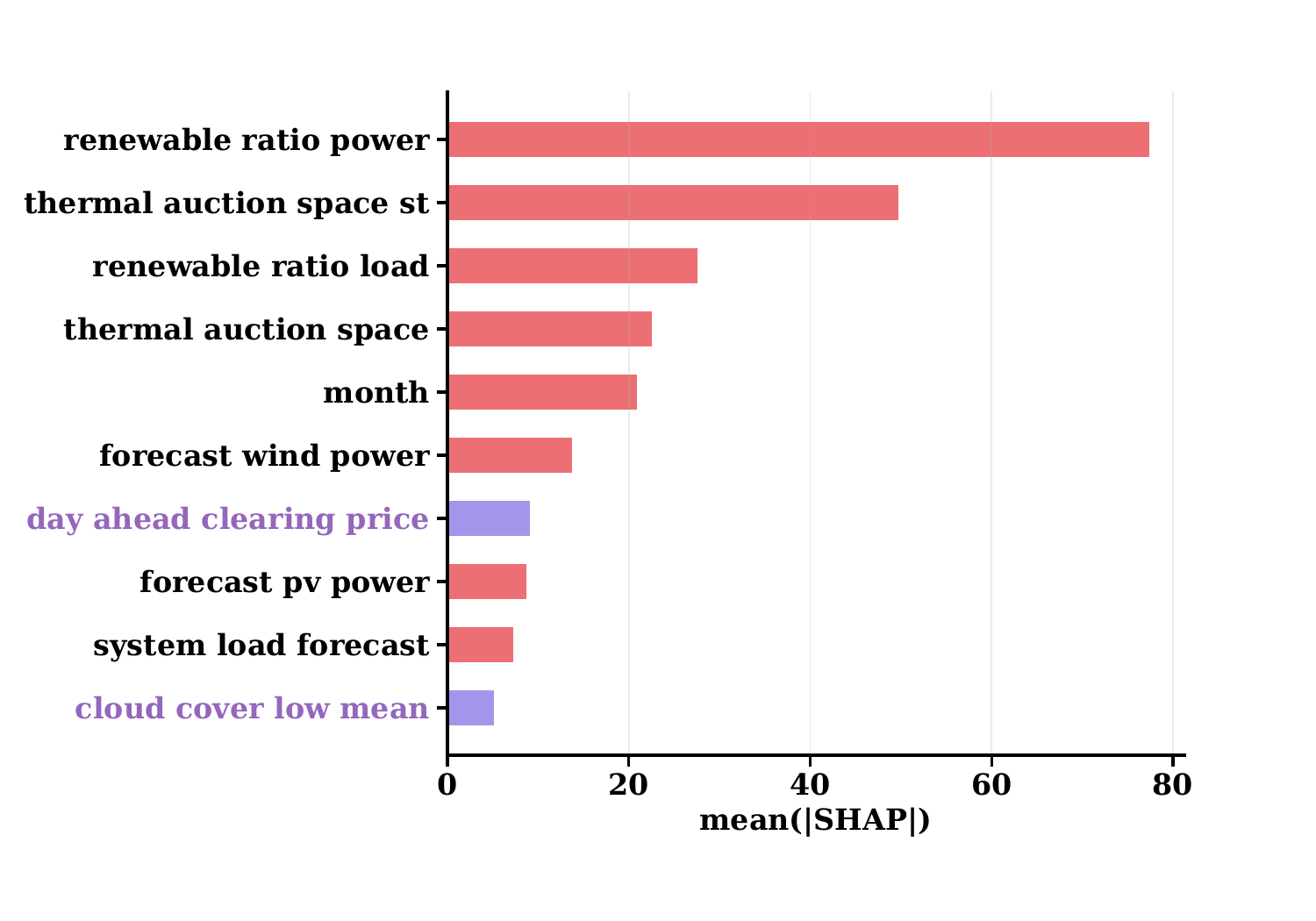}
    \caption{Global SHAP explanations on the test split. The Chronos2-derived forecast feature is highlighted to distinguish TSFM-predicted drivers from other exogenous variables.}
    \label{fig:shap_global_bar}
\end{figure}

\textbf{Global explanations.}
We compute SHAP values on the held-out test split in the rolling evaluation and summarize the results with two complementary global views in Fig.~\ref{fig:shap_global} and Fig.~\ref{fig:shap_global_bar}.

Fig.~\ref{fig:shap_global} presents a beeswarm distribution for the top features, where each point corresponds to one test sample and the horizontal axis indicates the signed SHAP contribution to the forecast.
The color encodes the raw feature value from low to high, revealing how feature magnitude relates to positive or negative impacts on the model output.

Fig.~\ref{fig:shap_global_bar} reports the mean absolute SHAP values, providing an overall importance ranking across all test samples.
Across both views, the most influential drivers are renewable-ratio variables and thermal-auction-space indicators, indicating that renewable availability and supply tightness play central roles in the regressor corrections.
The Chronos2-derived day-ahead clearing price forecast feature is highlighted to distinguish TSFM-derived signals from other exogenous variables, and it consistently appears among the top drivers, suggesting that the final model leverages the TSFM forecast while still relying substantially on domain exogenous information.

\textbf{Case studies.}
Fig.~\ref{fig:shap_cases_low_high} shows two representative deployed instances that illustrate how FutureBoosting improves extreme-regime forecasting beyond the Chronos2-only baseline.
Each instance is presented with two aligned views.
The left view is a SHAP waterfall for the selected step marked by the dashed vertical line in the waveform.
The right view plots the full trajectory over the same instance, comparing the ground truth, the Chronos2 forecast, and the FutureBoosting forecast.
In the waterfall, the model output is expressed as a global baseline plus feature-wise SHAP contributions, where positive values increase the forecast and negative values decrease it.
The TSFM-based feature derived from Chronos2 is highlighted to emphasize how the downstream regressor adjusts the base TSFM forecast using exogenous drivers.

In the low-price regime shown in Fig.~\ref{fig:case_low}, the true price drops sharply while Chronos2 remains biased high.
FutureBoosting follows the trough more closely, indicating systematic downward corrections learned in the enriched feature space.
The corresponding waterfall attributes the correction primarily to large negative contributions from renewable ratio power, month, thermal auction space st, and renewable ratio load, with additional negative effects from thermal auction space and other supporting variables.
These attributions suggest that renewable-availability signals and tightness-related indicators jointly drive the mitigation of overestimation during extreme low-price events.

In the high-price regime shown in Fig.~\ref{fig:case_high}, the true price exhibits a pronounced spike that is underestimated by Chronos2.
FutureBoosting amplifies the forecast and better matches the observed surge.
The waterfall assigns most of the upward adjustment to thermal auction space and thermal auction space st, together with positive contributions from renewable ratio power and other supply-demand drivers.

\begin{figure}[htbp]
  \centering
  \begin{subfigure}[t]{0.95\linewidth}
    \centering
    \includegraphics[width=\linewidth]{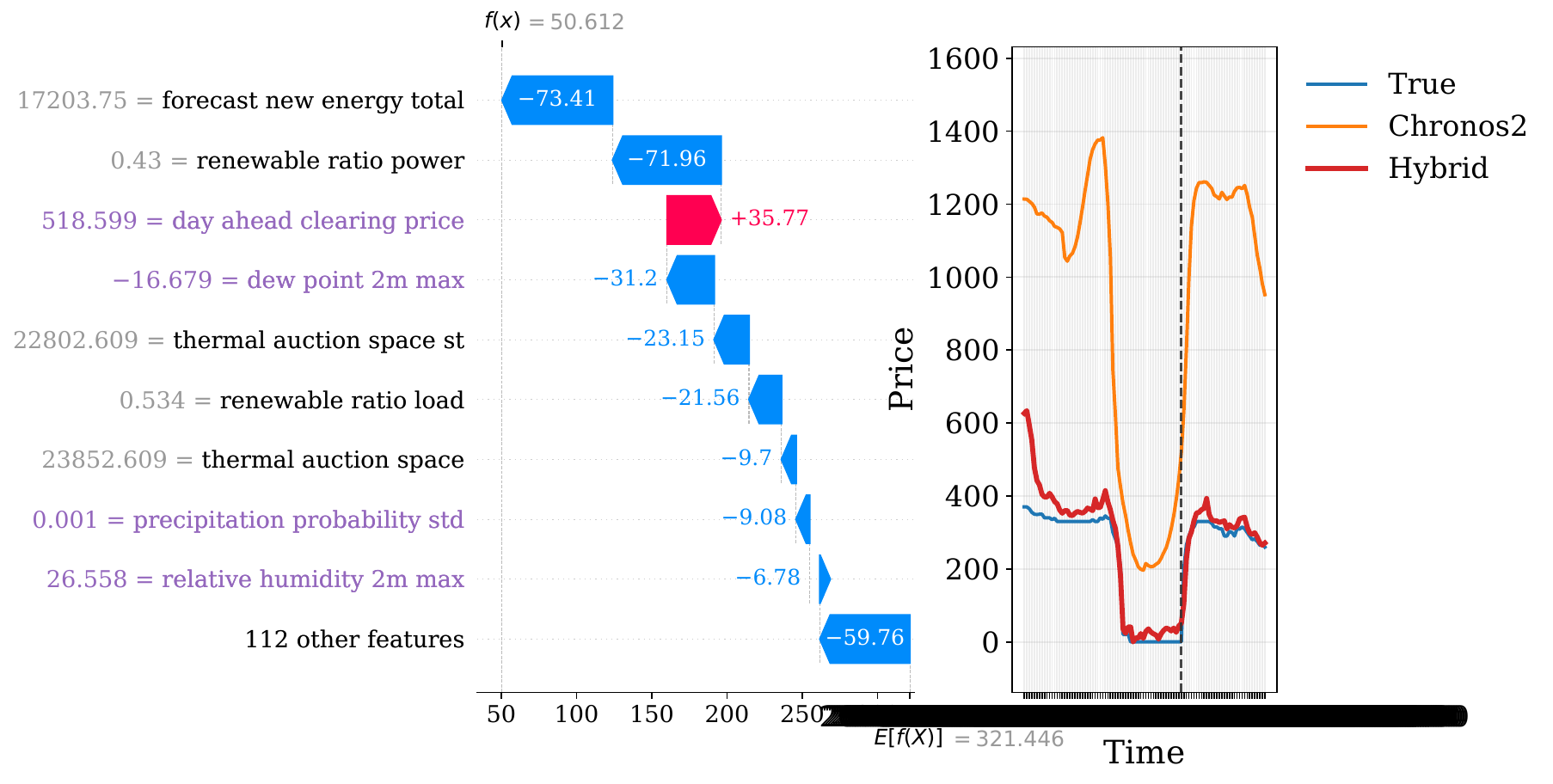}
    \caption{A low-price case in January 2025.}
    \label{fig:case_low_1}
  \end{subfigure}
  \begin{subfigure}[t]{0.95\linewidth}
    \centering
    \includegraphics[width=\linewidth]{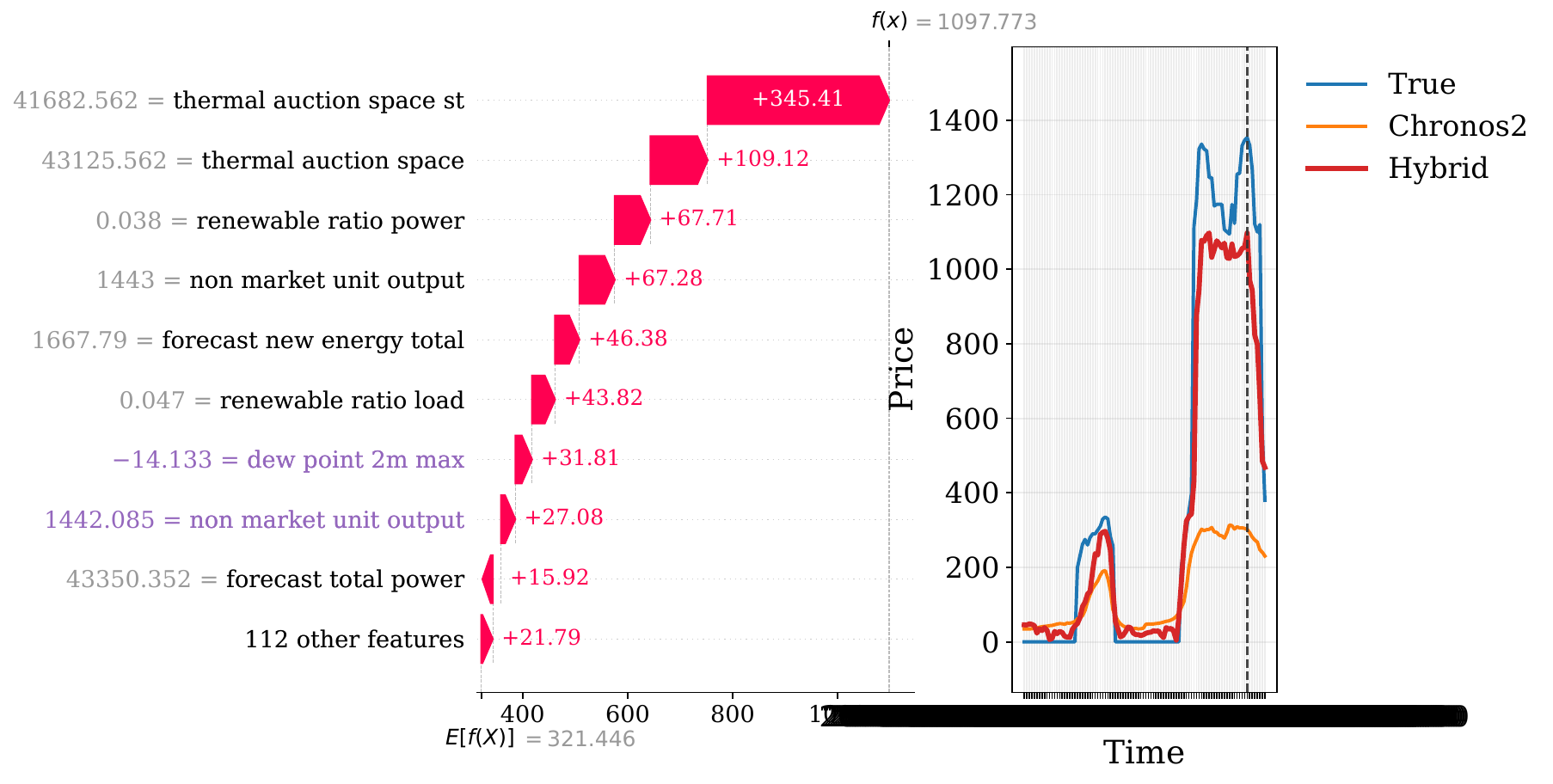}
    \caption{A high-price case in January 2025.}
    \label{fig:case_high_1}
  \end{subfigure}

  \caption{Additional sample SHAP explanations on deployed data. The TSFM-based feature derived from Chronos2 is highlighted.}
  \label{fig:shap_cases_1}
\end{figure}

To further complement the qualitative evidence, we provide two additional representative instances in Fig.~\ref{fig:shap_cases_1}, covering one low-price and one high-price case from January 2025.

\section{Month-wise Results on Shanxi}
\label{sec:appendix_shanxi_monthly}

Tables~\ref{tab:app_shanxi_12m_zs_hyb} and~\ref{tab:app_shanxi_12m_rt_zs_hyb} provide the detailed
month-wise results for the 12 rolling test windows in 2025.
For each TSFM, we report the zero-shot forecast (ZS) and its corresponding FutureBoosting regressor (FBR) for every month,
where lower values indicate better performance. Each month is listed with two rows (ZS/FBR) to facilitate
a direct within-window comparison.

\begin{table*}[t]
\centering
\caption{Shanxi Day-Ahead 12-month window-wise results comparing zero-shot (ZS) TSFM-only forecasts and their corresponding FutureBoosting (FBR).}
\label{tab:app_shanxi_12m_zs_hyb}
\small
\setlength{\tabcolsep}{1.6pt}
\renewcommand{\arraystretch}{1.05}

\begin{tabular*}{\textwidth}{@{\extracolsep{\fill}} l c *{14}{c} @{}}
\toprule
\multirow{2}{*}{Window} & \multirow{2}{*}{Type} &
\multicolumn{2}{c}{TimerXL} &
\multicolumn{2}{c}{Sundial} &
\multicolumn{2}{c}{Chronos2 (mv)} &
\multicolumn{2}{c}{Chronos2 (future)} &
\multicolumn{2}{c}{TimesFM} &
\multicolumn{2}{c}{Moirai2} &
\multicolumn{2}{c}{TiRex} \\
\cmidrule(lr){3-4}\cmidrule(lr){5-6}\cmidrule(lr){7-8}\cmidrule(lr){9-10}
\cmidrule(lr){11-12}\cmidrule(lr){13-14}\cmidrule(lr){15-16}
& & MSE & MAE & MSE & MAE & MSE & MAE & MSE & MAE & MSE & MAE & MSE & MAE & MSE & MAE \\
\midrule

\multirow{2}{*}{202501} & ZS  & 96031.92 & 195.56 & 77180.33 & 163.32 & 89610.94 & 175.98 & 66888.00 & 159.64 & 73564.20 & 160.36 & 81880.01 & 166.92 & 79950.36 & 161.65 \\
 & FBR & 28336.35 & 89.36 & 21958.62 & 84.95 & 18692.07 & 77.39 & 20340.07 & 87.74 & 20774.21 & 79.08 & 19424.70 & 80.66 & 20981.86 & 83.95 \\
  \cmidrule(lr){3-16}

\multirow{2}{*}{202502} & ZS  & 134084.19 & 237.77 & 121219.58 & 220.57 & 114767.00 & 220.21 & 92523.61 & 211.76 & 126918.32 & 213.88 & 123831.87 & 222.76 & 121434.01 & 210.61 \\
 & FBR & 40757.15 & 110.22 & 45136.53 & 123.09 & 44370.32 & 120.87 & 48430.27 & 125.53 & 44246.05 & 116.91 & 48259.19 & 119.29 & 43017.85 & 118.11 \\
  \cmidrule(lr){3-16}

\multirow{2}{*}{202503} & ZS  & 93755.35 & 196.24 & 80372.27 & 185.39 & 78314.09 & 172.93 & 66959.59 & 161.66 & 97939.16 & 180.96 & 90217.22 & 173.04 & 95215.36 & 165.77 \\
 & FBR & 45468.58 & 109.31 & 42478.45 & 111.75 & 43072.86 & 109.83 & 46726.38 & 110.72 & 42435.10 & 110.51 & 48687.06 & 115.28 & 45406.77 & 115.15 \\
  \cmidrule(lr){3-16}

\multirow{2}{*}{202504} & ZS  & 77669.32 & 175.90 & 57050.10 & 141.62 & 53054.38 & 137.56 & 41129.56 & 122.26 & 60371.87 & 133.42 & 58169.98 & 134.34 & 62035.25 & 128.32 \\
 & FBR & 26432.77 & 79.77 & 29912.20 & 83.65 & 30530.24 & 83.57 & 31365.62 & 84.79 & 33709.88 & 86.42 & 29063.96 & 83.90 & 34297.45 & 88.75 \\
  \cmidrule(lr){3-16}

\multirow{2}{*}{202505} & ZS  & 93666.34 & 180.71 & 86903.53 & 172.03 & 100305.20 & 185.01 & 92857.96 & 192.45 & 86284.90 & 162.18 & 90186.08 & 166.77 & 87845.80 & 153.05 \\
 & FBR & 32690.44 & 92.38 & 37007.00 & 97.54 & 37003.55 & 98.47 & 36565.64 & 96.68 & 35779.71 & 96.44 & 31915.10 & 90.78 & 36821.45 & 97.27 \\
  \cmidrule(lr){3-16}

\multirow{2}{*}{202506} & ZS  & 48227.93 & 128.14 & 25582.65 & 95.52 & 27434.78 & 92.98 & 24986.22 & 88.57 & 30333.88 & 94.76 & 30058.37 & 91.84 & 33013.53 & 88.16 \\
 & FBR & 21909.27 & 86.89 & 23753.56 & 86.64 & 24089.27 & 85.46 & 21861.27 & 76.54 & 20864.53 & 77.29 & 20510.49 & 77.03 & 22181.49 & 82.93 \\
  \cmidrule(lr){3-16}

\multirow{2}{*}{202507} & ZS  & 47276.94 & 113.89 & 34349.32 & 94.85 & 35194.34 & 103.91 & 24992.95 & 97.87 & 37760.89 & 93.24 & 41461.06 & 96.68 & 45637.71 & 98.15 \\
 & FBR & 22805.03 & 85.41 & 28215.34 & 90.66 & 24555.22 & 87.92 & 32517.70 & 96.25 & 25388.54 & 86.02 & 24899.19 & 86.08 & 26646.30 & 88.50 \\
  \cmidrule(lr){3-16}

\multirow{2}{*}{202508} & ZS  & 26194.18 & 92.84 & 18446.30 & 84.86 & 19041.06 & 79.95 & 22911.02 & 82.82 & 21042.63 & 82.44 & 16232.81 & 70.61 & 19955.39 & 73.31 \\
 & FBR & 31828.17 & 90.12 & 25673.99 & 77.76 & 28298.30 & 81.09 & 24676.54 & 84.59 & 31304.71 & 83.44 & 35594.87 & 92.77 & 30230.98 & 85.11 \\
  \cmidrule(lr){3-16}

\multirow{2}{*}{202509} & ZS  & 58305.51 & 124.13 & 50975.25 & 115.60 & 54196.83 & 128.87 & 48346.62 & 120.97 & 55041.01 & 122.43 & 54495.37 & 118.30 & 54019.00 & 113.74 \\
 & FBR & 32633.16 & 75.44 & 39071.02 & 80.19 & 38111.77 & 80.32 & 31227.95 & 75.61 & 40301.52 & 81.39 & 37861.71 & 80.32 & 41216.20 & 83.26 \\
  \cmidrule(lr){3-16}

\multirow{2}{*}{202510} & ZS  & 106548.72 & 218.57 & 88226.08 & 194.62 & 77523.35 & 183.72 & 54935.98 & 161.40 & 91218.39 & 188.55 & 87374.95 & 182.83 & 90006.81 & 185.27 \\
 & FBR & 79067.78 & 167.47 & 85247.65 & 173.37 & 81003.48 & 169.03 & 84240.63 & 175.93 & 87127.87 & 176.08 & 73088.76 & 163.71 & 81226.46 & 167.61 \\
  \cmidrule(lr){3-16}

\multirow{2}{*}{202511} & ZS  & 33343.31 & 111.37 & 27710.86 & 100.95 & 27223.86 & 97.97 & 22383.87 & 86.41 & 27917.56 & 90.34 & 25812.17 & 82.76 & 26125.74 & 75.93 \\
 & FBR & 11194.69 & 55.77 & 9789.58 & 48.10 & 8740.34 & 49.97 & 11148.39 & 52.26 & 8581.90 & 47.50 & 11883.00 & 54.63 & 8719.66 & 48.81 \\
  \cmidrule(lr){3-16}

\multirow{2}{*}{202512} & ZS  & 100243.11 & 177.10 & 87738.88 & 153.87 & 83058.95 & 155.21 & 67451.35 & 158.16 & 88694.82 & 145.02 & 89289.72 & 144.10 & 94168.80 & 143.73 \\
 & FBR & 34617.97 & 95.26 & 38793.17 & 107.41 & 33662.70 & 94.84 & 34679.67 & 99.04 & 34428.32 & 99.38 & 37974.19 & 104.36 & 35774.00 & 103.80 \\

\midrule
\multirow{2}{*}{\textbf{AVG}} & ZS  & 76278.90 & 162.69 & 62979.60 & 143.60 & 63310.40 & 144.53 & 52197.23 & 137.00 & 66423.97 & 138.97 & 65750.80 & 137.58 & 67450.65 & 133.14 \\
 & FBR & 33978.45 & 94.78 & 35586.43 & 97.09 & 34344.18 & 94.90 & 35315.01 & 97.14 & 35411.86 & 95.04 & 34930.18 & 95.73 & 35543.37 & 96.94 \\
\cmidrule(lr){3-16}
\multicolumn{2}{l}{\textbf{Improve (\%)}} &
+55.45 & +41.74 &
+43.50 & +32.39 &
+45.75 & +34.34 &
+32.34 & +29.09 &
+46.69 & +31.61 &
+46.87 & +30.42 &
+47.30 & +27.19 \\
\bottomrule
\end{tabular*}
\end{table*}

\begin{table*}[t]
\centering
\caption{Shanxi Real-time 12-month window-wise results comparing zero-shot (ZS) TSFM-only forecasts and their corresponding FutureBoosting (FBR).}
\label{tab:app_shanxi_12m_rt_zs_hyb}
\small
\setlength{\tabcolsep}{1.6pt}
\renewcommand{\arraystretch}{1.05}

\begin{tabular*}{\textwidth}{@{\extracolsep{\fill}} l c *{14}{c} @{}}
\toprule
\multirow{2}{*}{Window} & \multirow{2}{*}{Type} &
\multicolumn{2}{c}{TimerXL} &
\multicolumn{2}{c}{Sundial} &
\multicolumn{2}{c}{Chronos2 (mv)} &
\multicolumn{2}{c}{Chronos2 (future)} &
\multicolumn{2}{c}{TimesFM} &
\multicolumn{2}{c}{Moirai2} &
\multicolumn{2}{c}{TiRex} \\
\cmidrule(lr){3-4}\cmidrule(lr){5-6}\cmidrule(lr){7-8}\cmidrule(lr){9-10}
\cmidrule(lr){11-12}\cmidrule(lr){13-14}\cmidrule(lr){15-16}
& & MSE & MAE & MSE & MAE & MSE & MAE & MSE & MAE & MSE & MAE & MSE & MAE & MSE & MAE \\
\midrule

\multirow{2}{*}{202501} & ZS  & 118895.21 & 217.91 & 97693.37 & 190.56 & 114730.62 & 199.96 & 99201.90 & 195.49 & 98345.31 & 188.98 & 114230.96 & 202.83 & 114310.87 & 198.35 \\
 & FBR & 23393.77  & 100.68 & 27793.07 & 107.87 & 28962.75  & 108.12 & 41628.91 & 127.75 & 31895.88 & 114.04 & 27384.44  & 106.69 & 34742.31  & 121.41 \\
  \cmidrule(lr){3-16}

\multirow{2}{*}{202502} & ZS  & 120379.08 & 229.16 & 85619.59 & 181.18 & 90050.68 & 203.80 & 80630.58 & 209.08 & 98546.03 & 191.38 & 117523.33 & 222.87 & 97648.07 & 179.53 \\
 & FBR & 32541.10  & 119.51 & 36259.24 & 126.50 & 38276.06  & 124.70 & 37569.89 & 123.42 & 32796.34 & 116.31 & 35840.88  & 115.25 & 38160.78  & 117.90 \\
  \cmidrule(lr){3-16}

\multirow{2}{*}{202503} & ZS  & 107854.53 & 206.39 & 95016.02 & 202.85 & 88038.66 & 191.04 & 77018.86 & 187.73 & 111121.30 & 196.46 & 102835.81 & 195.97 & 109298.64 & 178.06 \\
 & FBR & 65560.99  & 135.07 & 67539.49 & 135.37 & 63971.96  & 136.43 & 69542.25 & 137.31 & 64435.87 & 136.05 & 65943.71  & 137.51 & 68689.22  & 138.64 \\
  \cmidrule(lr){3-16}

\multirow{2}{*}{202504} & ZS  & 82330.52 & 182.88 & 67713.56 & 151.20 & 68948.49 & 168.73 & 64970.55 & 177.79 & 80565.65 & 153.57 & 75139.75 & 150.78 & 76841.29 & 140.66 \\
 & FBR & 38817.92 & 116.73 & 43164.55 & 121.84 & 38725.26 & 106.06 & 49110.38 & 120.86 & 41578.67 & 106.84 & 38528.82 & 108.33 & 41703.71 & 107.60 \\
  \cmidrule(lr){3-16}

\multirow{2}{*}{202505} & ZS  & 102967.35 & 188.26 & 89835.88 & 171.51 & 106919.77 & 212.81 & 108991.74 & 232.27 & 113641.81 & 184.17 & 97904.36 & 176.92 & 95134.62 & 155.97 \\
 & FBR & 45863.10  & 118.49 & 44887.32 & 117.20 & 49086.25  & 116.84 & 47795.90 & 121.09 & 48432.05 & 119.94 & 46761.93  & 122.08 & 53846.44  & 123.05 \\
  \cmidrule(lr){3-16}

\multirow{2}{*}{202506} & ZS  & 65641.39 & 139.82 & 47375.07 & 121.01 & 47714.17 & 120.61 & 44323.24 & 120.88 & 48809.47 & 109.90 & 49823.53 & 115.69 & 52145.87 & 112.20 \\
 & FBR & 42519.09 & 110.62 & 44953.33 & 119.60 & 43102.85 & 118.05 & 40441.20 & 107.20 & 40049.48 & 108.22 & 43203.69 & 114.87 & 41016.73 & 110.74 \\
  \cmidrule(lr){3-16}

\multirow{2}{*}{202507} & ZS  & 41142.87 & 108.49 & 30470.52 & 89.86 & 31539.87 & 99.73 & 27781.38 & 106.07 & 32773.53 & 87.15 & 41767.10 & 95.36 & 36157.71 & 87.12 \\
 & FBR & 25942.39  & 99.99 & 32145.62 & 99.94 & 22851.93  & 97.04 & 23634.30 & 92.29 & 25053.84 & 104.78 & 21023.51  & 93.36 & 26019.67  & 96.04 \\
  \cmidrule(lr){3-16}

\multirow{2}{*}{202508} & ZS  & 31030.77 & 94.68 & 25916.09 & 90.20 & 26063.24 & 92.93 & 29037.41 & 97.29 & 33563.30 & 94.70 & 23631.02 & 79.13 & 28671.56 & 80.56 \\
 & FBR & 42816.39 & 108.70 & 36533.89 & 100.93 & 43829.60 & 107.10 & 36597.41 & 105.57 & 40358.88 & 104.96 & 45826.41 & 107.22 & 43251.67 & 105.40 \\
  \cmidrule(lr){3-16}

\multirow{2}{*}{202509} & ZS  & 85633.09 & 156.08 & 78310.54 & 154.30 & 79575.40 & 165.89 & 76442.12 & 172.88 & 79838.22 & 146.30 & 86435.07 & 156.92 & 83710.00 & 146.00 \\
 & FBR & 57474.52 & 109.76 & 62770.34 & 110.22 & 63742.47 & 113.22 & 59888.76 & 114.83 & 64865.33 & 110.78 & 63983.58 & 113.40 & 63109.99 & 113.77 \\
  \cmidrule(lr){3-16}

\multirow{2}{*}{202510} & ZS  & 143685.32 & 245.75 & 125149.12 & 234.85 & 91568.20 & 210.20 & 66799.12 & 180.18 & 155995.52 & 259.92 & 127427.50 & 228.15 & 134176.60 & 232.44 \\
 & FBR & 112637.92 & 203.07 & 105293.65 & 196.79 & 120206.91 & 215.92 & 102746.86 & 200.65 & 122923.08 & 214.33 & 115473.27 & 205.10 & 110535.59 & 199.50 \\
  \cmidrule(lr){3-16}

\multirow{2}{*}{202511} & ZS  & 38809.75 & 120.18 & 36503.10 & 114.82 & 31784.27 & 107.45 & 28197.25 & 102.97 & 34545.62 & 99.15 & 31778.31 & 96.23 & 32763.18 & 87.15 \\
 & FBR & 16096.67 & 73.50 & 16402.15 & 74.61 & 16514.01 & 75.48 & 17205.48 & 71.43 & 16239.71 & 72.56 & 16502.11 & 76.04 & 15307.39 & 71.51 \\
  \cmidrule(lr){3-16}

\multirow{2}{*}{202512} & ZS  & 97874.73 & 169.98 & 87027.78 & 149.20 & 93279.61 & 155.96 & 74450.43 & 171.96 & 91786.17 & 148.05 & 89730.76 & 145.00 & 88617.30 & 129.12 \\
 & FBR & 30602.62 & 109.11 & 27243.63 & 98.49 & 31096.27 & 108.19 & 34042.47 & 112.00 & 32710.28 & 107.72 & 31379.47 & 112.55 & 26954.77 & 102.59 \\

\midrule
\multirow{2}{*}{\textbf{AVG}} & ZS  & 86353.72 & 171.63 & 72219.22 & 154.30 & 72517.75 & 160.76 & 64820.38 & 162.88 & 81627.66 & 154.98 & 79852.29 & 155.49 & 79122.98 & 143.93 \\
 & FBR & 44522.21 & 117.10 & 45415.52 & 117.45 & 46697.19 & 118.93 & 46683.65 & 119.53 & 46778.28 & 118.04 & 45987.65 & 117.70 & 46944.86 & 117.35 \\
\cmidrule(lr){3-16}
\multicolumn{2}{l}{\textbf{Improve (\%)}} &
+48.44 & +31.77 &
+37.11 & +23.88 &
+35.61 & +26.02 &
+27.98 & +26.61 &
+42.69 & +23.83 &
+42.41 & +24.30 &
+40.67 & +18.47 \\
\bottomrule
\end{tabular*}
\end{table*}

\end{document}